# Hybrid tractability of
# soft constraint problems[*]

Martin C. Cooper
IRIT, University of Toulouse III
cooper@irit.fr

Stanislav Živný[†]
Oxford University
standa.zivny@comlab.ox.ac.uk

**Abstract**

The constraint satisfaction problem (CSP) is a central generic problem in computer science and artificial intelligence: it provides a common framework for many theoretical problems as well as for many real-life applications. Soft constraint problems are a generalisation of the CSP which allow the user to model optimisation problems. Considerable effort has been made in identifying properties which ensure tractability in such problems. In this work, we initiate the study of hybrid tractability of soft constraint problems; that is, properties which guarantee tractability of the given soft constraint problem, but which do not depend only on the underlying structure of the instance (such as being tree-structured) or only on the types of soft constraints in the instance (such as submodularity). We present several novel hybrid classes of soft constraint problems, which include a machine scheduling problem, constraint problems of arbitrary arities with no overlapping nogoods, and the SOFTALLDIFF constraint with arbitrary unary soft constraints. An important tool in our investigation will be the notion of forbidden substructures.

## 1 Introduction

An instance of the constraint satisfaction problem (CSP) consists of a collection of variables which must be assigned values subject to specified constraints. Each CSP instance has an underlying undirected graph, known as its *constraint network*, whose vertices are the variables of the instance, and two vertices are adjacent if corresponding variables are related by some constraint. Such a graph is also known as the *structure* of the instance.

An important line of research on the CSP is to identify all tractable cases which are recognisable in polynomial time. Most of this work has been focused on one of the two general approaches: either identifying forms of constraint which are sufficiently restrictive to ensure tractability no matter how they are combined [1, 2], or else identifying structural properties of constraint networks which ensure tractability no matter what forms of constraint are imposed [3].

The first approach has led to identifying certain algebraic properties known as polymorphisms [4] which are necessary for a set of constraint types to ensure tractability. A set of constraint types which ensures tractability is called a *tractable constraint language*. The second approach has been used to characterise all tractable cases of bounded-arity CSPs (such as binary CSPs): the *only* class of structures which ensures tractability (subject to certain complexity theory assumptions) are structures of *bounded tree-width* [5, 6].

In practice, constraint satisfaction problems usually do not possess a sufficiently restricted structure or use a sufficiently restricted constraint language to fall into any of these tractable classes. Nevertheless, they may still have properties which ensure they can be solved efficiently, but these properties concern both the structure and the form of the constraints. Such properties have sometimes been called *hybrid* reasons for tractability [7, 8, 9, 10, 11].

---

[*]A preliminary version of part of this work appeared in *Proceedings of the 16th International Conference on Principles and Practice of Constraint Programming (CP)*, pp. 152–166, LNCS 6308, 2010.

[†]Stanislav Živný acknowledges the support of EPSRC grant EP/F01161X/1, EPSRC PhD+ Award, and Junior Research Fellowship at University College, Oxford.

Since in practice many constraint satisfaction problems are over-constrained, and hence have no solution, or are under-constrained, and hence have many solutions, *soft* constraint satisfaction problems have been studied [7]. In an instance of the soft CSP, every constraint is associated with a function (rather than a relation as in the CSP) which represents preferences among different partial assignments, and the goal is to find the best assignment. Several very general soft CSP frameworks have been proposed in the literature [12, 13]. In this paper we focus on one of the very general frameworks, the *valued* constraint satisfaction problem (VCSP) [12].

Similarly to the CSP, an important line of research on the VCSP is to identify tractable cases which are recognisable in polynomial time. Is is well known that structural reasons for tractability generalise to the VCSP [7]. In the case of language restrictions, only a few conditions are known to guarantee tractability of a given set of valued constraints [14, 15].

Up until now there have been very few results on hybrid tractability for the VCSP. For instance, Kumar defines an interesting framework for hybrid tractability for the Boolean weighted CSP [10]. However, to the best of our knowledge, this framework has so far not provided any new hybrid classes. In fact, all tractable classes presented in [10] are not hybrid and are already known.

**Contributions** The main contribution of the paper is the systematic study of hybrid tractability of VCSPs and the introduction of several novel hybrid tractable classes of VCSPs. As a first step, we start with binary VCSPs.

First, we will demonstrate three hybrid classes defined by forbidding certain graphs as induced subgraphs in the structure of the VCSP instance.[1] However, the tractable classes described in this part of the paper are not entirely satisfactory as a first step towards a general theory of hybrid tractable classes of VCSP instances, since the only soft constraints they allow are unary and, furthermore, classical graph theory does not even allow us to express all possible classes of such instances.

Second, we introduce classes of CSPs with polynomially many solutions and polynomially many dead ends in the search tree. We also discuss tractable extensions of these two classes to VCSPs. Furthermore, we show that the recently introduced hybrid class of CSPs satisfying the broken-triangle property [11] is not extendible to VCSPs with soft unary constraints.

Finally, as arguably the most interesting hybrid class in this paper, we introduce the class defined by the *joint-winner property* (JWP). This class generalises the SOFTALLDIFF constraint with arbitrary unary soft constraints. Moreover, this class can be generalised to a larger class of non-binary VCSPs which includes CSP instances with no overlapping nogoods.

A preliminary version of part of this work appeared as a conference paper [16]. This paper is a significant extension of the previous work [16] since it includes: an extended Section 3 (VCSPs with crisp binary constraints), new hybrid tractable classes (based on polynomially many solutions and dead ends, described in Section 4), an improved complexity analysis of the algorithm solving VCSP instances satisfying the JWP in Section 5.3, a complete proof of maximality of the JWP in Section 5.4 ([16] considered only a special case), and an extension of the JWP in Section 5.5. Moreover, the preliminary conference version [16] did not consider the *Z-free* property, discussed in Section 5.2, which is necessary for the complete proof of tractability of VCSPs satisfying the JWP.

The rest of the paper is organised as follows. In Section 2, we define binary constraint satisfaction problems (CSPs), valued constraint satisfaction problems (VCSPs) and other necessary definitions needed throughout the paper. In Section 3, we study binary VCSPs whose only soft constraints are unary. Using a connection between these VCSPs and the maximum weight independent set problem in certain graph classes, we identify hybrid tractable classes of VCSPs. In Section 4, we present a novel class of tractable CSPs based on CSPs with polynomially many dead ends in the search tree. Furthermore, we present a class of tractable CSPs with polynomially many solutions. We discuss tractable extensions of these two classes to VCSPs. In Section 5, we define the joint-winner property. In Section 5.1, we give several examples of studied problems that satisfy the joint-winner property. In Section 5.2, we study important properties of VCSP instances satisfying the joint-winner property, which allow us, in Section 5.3, to present a polynomial-time algorithm for solving binary VCSPs satisfying the joint-winner property. In Section 5.4, we prove that this new tractable class is maximal. In Section 5.5, we extend the class of tractable VCSPs defined by the joint-winner property. Finally, in Section 6, we summarise our work and finish with some open problems.

---
[1]More precisely, in the *micro-structure complement* of the instance (defined in Section 2).

We remark that even though our results are formulated as soft constraint satisfaction problems, it is clear that these results apply to various other optimisation frameworks that are equivalent to soft constraint problems such as Gibbs energy minimisation, Markov Random Fields and other graphical models [17, 18].

## 2 Preliminaries

In this paper we focus on binary valued constraint satisfaction problems. We denote by $\mathbb{Q}_+$ the set of all non-negative rational numbers. We denote $\overline{\mathbb{Q}}_+ = \mathbb{Q}_+ \cup \{\infty\}$ with the standard addition operation extended so that for all $a \in \overline{\mathbb{Q}}_+$, $a + \infty = \infty$. Members of $\overline{\mathbb{Q}}_+$ are called *costs*.

A unary cost function over domain $D_i$ is a mapping $c_i : D_i \to \overline{\mathbb{Q}}_+$. A binary cost function over domains $D_i$ and $D_j$ is a mapping $c_{ij} : D_i \times D_j \to \overline{\mathbb{Q}}_+$. If the range of $c_i$ ($c_{ij}$ respectively) lies entirely within $\mathbb{Q}_+$, then $c_i$ ($c_{ij}$ respectively) is called a *finite-valued* cost function.

If the range of $c_i$ ($c_{ij}$ respectively) is $\{\alpha, \infty\}$, for some $\alpha \in \mathbb{Q}_+$, then $c_i$ ($c_{ij}$ respectively) is called a *crisp* cost function. Note that crisp cost functions are just relations; that is, subsets of $D_i$ (in the unary case) or $D_i \times D_j$ (in the binary case) corresponding to the set of finite-cost tuples. If $c_i$ ($c_{ij}$ respectively) is not a crisp cost function, it is called *soft*.

A binary VCSP instance [12] consists of a set of *variables* (denoted as $v_i$, where $i \in \{1, \ldots, n\}$); for each variable $v_i$ a *domain* $D_i$ containing possible *values* for variable $v_i$; and a set of *valued constraints*. Each valued constraint is either of the form $\langle v_i, c_i \rangle$, where $v_i$ is a variable and $c_i$ is a unary cost function (constraints of this form are called *unary* constraints), or of the form $\langle \langle v_i, v_j \rangle, c_{ij} \rangle$, where $v_i$ and $v_j$ are variables, the pair $\langle v_i, v_j \rangle$ is called the *scope* of the constraint, and $c_{ij}$ is a binary cost function (constraints of this form are called *binary* constraints). A constraint is called crisp if its associated cost function is crisp, and similarly a constraint is called soft if its associated cost function is soft. For notational convenience, throughout this paper we assume that there is a unique valued constraint on any given scope. In particular, $c_{ij}(a, b) = c_{ji}(b, a)$, since they are simply two different ways of representing the unique cost of simultaneously assigning $\langle a, b \rangle$ to variables $\langle i, j \rangle$.

A *solution* to a VCSP instance is an assignment of values from the domains to the variables with the minimum total cost given by

$$\sum_{i=1}^{n} c_i(v_i) + \sum_{1 \leq i < j \leq n} c_{ij}(v_i, v_j).$$

A VCSP instance with only crisp constraints is called a CSP instance. In the CSP, the task of finding an optimal solution amounts to deciding whether there is a solution with finite cost (all constraints are satisfied).

The *micro-structure* of a binary CSP instance $\mathcal{P}$ is a graph where the set of vertices corresponds to the set of possible assignments of values to variables: a vertex $\langle v_i, a \rangle$ represents the assignment of value $a$ to variable $v_i$ [19]. The edges of the micro-structure connect all pairs of variable-value assignments that are allowed by the constraints. (Note that if there is no explicit constraint between two variables, then it is considered to be the complete constraint allowing all pairs of assignments.) The micro-structure of a binary VCSP instance is defined similarly, but both vertices and edges of the graph are assigned costs. For CSPs, the *micro-structure complement* is the complement of the micro-structure: its edges represent pairs of variable-value assignments that are disallowed by the constraints. Hence for every variable $v_i$, the micro-structure complement contains all edges of the form $\{\langle v_i, a \rangle, \langle v_i, b \rangle\}$ for $a \neq b \in D_i$ as every variable can be assigned only one value.

A *clique* in a graph is a set of vertices which are pairwise adjacent. An *independent* set in a graph is a set of vertices which are pairwise non-adjacent. It is well known that solving a CSP instance $\mathcal{P}$ is equivalent to finding a clique of size $n$ in the micro-structure of $\mathcal{P}$, and to finding an independent set of size $n$ in the micro-structure complement of $\mathcal{P}$ [19]. Therefore, tractability results on the maximum independent set problem for various classes of graphs can be straightforwardly used to obtain tractable CSP classes [8, 20].

Given a graph $G$, we denote by $V(G)$ the set of vertices of $G$ and by $E(G)$ the set of edges of $G$. A *coloured graph* $\mathcal{G} = \langle G, c_G \rangle$ is a graph $G$ and a colouring $c_G : V(G) \to \{1, \ldots, n\}$ of the vertices of $G$. The *coloured micro-structure (complement)* of an $n$-variable binary CSP instance $\mathcal{P}$ is $\mathcal{G} = \langle G, c_G \rangle$ where $G$ is the micro-structure (complement) of $\mathcal{P}$ and $c_G$ is a colouring $c_G : V(G) \to \{1, \ldots, n\}$ of the vertices of $G$, such that the colour of vertex $\langle v_i, a \rangle$ is $i$. Unlike the micro-structure, the coloured micro-structure contains all the information necessary to reconstruct the original instance $\mathcal{P}$.

A graph $H$ is an *induced subgraph* of $G$ if there is an injective mapping $f : V(H) \to V(G)$ such that $(u, v) \in E(H)$ if, and only if, $(f(u), f(v)) \in E(G)$. We now extend the notion of induced subgraphs to that of induced and forbidden substructures of the coloured micro-structure and coloured micro-structure complement. Let $\mathcal{G} = \langle G, c_G \rangle$ be the coloured micro-structure or coloured micro-structure complement of a CSP instance $\mathcal{P}$. Now let $\mathcal{H} = \langle H, c_H \rangle$ be a coloured graph. We say that $\mathcal{H}$ is an *induced substructure* of $\mathcal{G}$ if $H$ is an induced subgraph of $G$ and vertices of the same colour are mapped to vertices of the same colour; that is, if there is an injective mapping $f : V(H) \to V(G)$ such that (1) $(u, v) \in E(H)$ if and only if $(f(u), f(v)) \in E(G)$; and (2) $c_H(u) = c_H(v)$ if and only if $c_G(f(u)) = c_G(f(v))$. Throughout the paper we will present several classes of VCSPs that will be defined by *forbidding* a certain substructure (called a *pattern*) in the coloured micro-structure or the coloured micro-structure complement of the crisp part of the VCSP instance. In figures, we will often draw ovals around vertices that are assigned the same colour (corresponding to the assignments to the same variable).

## 3  VCSPs with crisp binary constraints

In this section we restrict our attention to binary VCSP instances with crisp binary constraints. There are no restrictions on unary constraints; hence both crisp and soft unary constraints are allowed. It follows from a recent result of Takhanov [33] that there is a P/NP-hard dichotomy for VCSPs in which all unary constraints are allowed and all binary crisp constraints belong to a particular language. It is an open question whether a similar dichotomy exists for hybrid classes. We show how tractability results on the maximum weight independent set in perfect, fork-free and apple-free graphs can be used to obtain hybrid tractable classes of VCSPs.

Let G be a graph $G = \langle V, E \rangle$ with weights $w : V \to \mathbb{Q}_+$ on the vertices of $G$. The weight of an independent set $S$ in $G$, denoted $w(S)$, is the sum of weights of the vertices in $S$: $w(S) = \sum_{v \in S} w(v)$. It is easy to show that given a binary VCSP instance $\mathcal{P}$ where only unary constraints can be soft, $\mathcal{P}$ can be solved by finding a maximum weight independent set in the micro-structure complement of $\mathcal{P}$ with weights given by $w(\langle v_i, a \rangle) = Mn - c_i(a)$, where $M$ is strictly greater than the maximum finite unary cost $c_i(a)$. Indeed, independent sets of weight strictly greater than $Mn(n-1)$ are in one-to-one correspondence with consistent assignments to $n$ variables in $\mathcal{P}$ (since $\forall i, \forall a, 0 \leq c_i(a) < M$ implies that the weight of an independent set of size $n$ is strictly greater than $n(Mn - M) = Mn(n-1)$ and the weight of an independent set of size $n-1$ is at most $Mn(n-1)$).

We now present several important and well-studied classes of graphs which admit a polynomial-time algorithm for the maximum weight independent set problem (MWIS).

Given a graph $G$, the *line graph* of $G$, denote by $L(G)$, is a graph such that each vertex of $L(G)$ represents an edge of $G$, and two vertices of $L(G)$ are adjacent if, and only if, their corresponding edges share a vertex in $G$. Beinike gave a characterisation of line graphs in terms of 9 forbidden induced subgraphs [21] (see Figure 1). Since an independent set in the line graph $L(G)$ of a graph $G$ corresponds to a matching of $G$, Edmond's algorithm for the maximum matching problem [22, 23] can be used for MWIS in line graphs.

A graph is *claw-free* if it does not contain a claw as an induced subgraph, where a claw is a complete bipartite graph $K_{1,3}$ with 1 vertex in one group and 3 vertices in the other group (see Figure 2 (b)). Since Beineke's characterisation of line graphs in terms of the set of nine forbidden induced subgraphs shown in Figure 1 includes a claw (top left graph in Figure 1), it follows that line graphs form a strict subset of claw-free graphs. Extending Edmond's algorithm for maximum matchings in graphs [22, 23], Minty designed

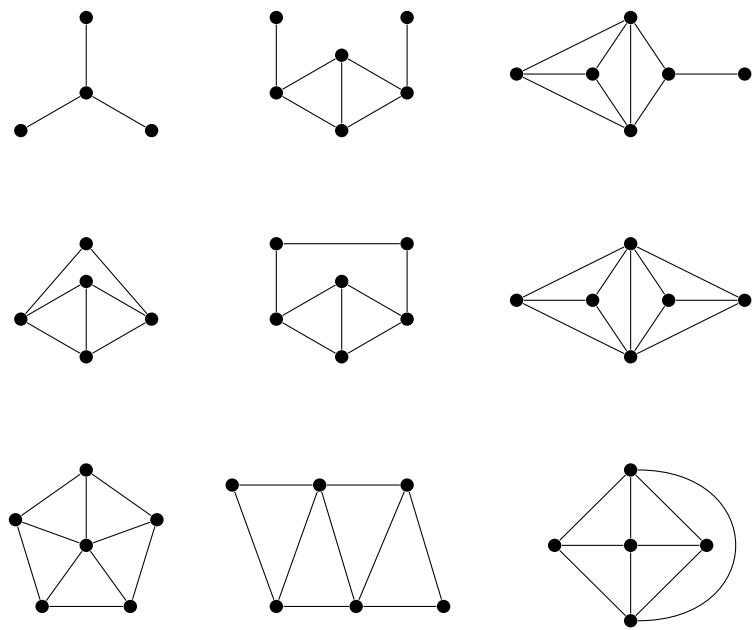

Figure 1: The nine minimal non-line graphs.

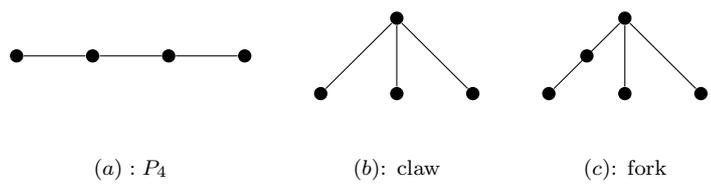

($a$) : $P_4$  ($b$): claw  ($c$): fork

Figure 2: A $P_4$, a claw and a fork.

an algorithm for the independent set problem in claw-free graphs [24]. Minty's algorithm was later corrected and extended to the maximum weight independent set problem in claw-free graphs [25].

A graph is $P_4$-*free* if it does not contain a $P_4$ as an induced subgraph, where $P_4$ is a chordless path on 4 vertices (see Figure 2 (a)). MWIS is known to be solvable in polynomial time in $P_4$-free graphs [26].

A graph is *fork-free* if it does not contain a fork (also called a chair) as an induced subgraph, where a fork is obtained from a claw by a single subdivision of its edges (see Figure 2 (c)). Since $P_4$ and claw are subgraphs of fork, it is clear that the class of fork-free graphs properly includes both $P_4$-free and claw-free graphs.

Lozin and Milanič have recently shown that MWIS is solvable in polynomial time in fork-free graphs [27], thus generalising the result of Alekseev for the unweighted case [28].

A graph is *apple-free* if it contains no $A_k$, $k \geq 4$, as an induced subgraph, where $A_k$, called an apple, is a graph obtained from a chordless cycle $C_k$ of length $k \geq 4$ by adding a vertex that has exactly one neighbour on the cycle (see Figure 3 for $A_4, A_5$ and $A_6$). The class of apple-free graphs properly includes claw-free graphs, and the MWIS problem is solvable in polynomial time [29].[2]

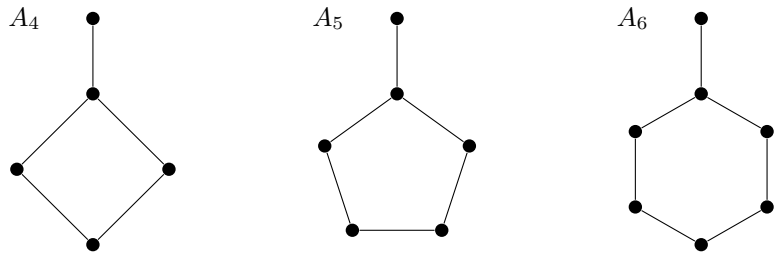

Figure 3: Apples.

We now turn our attention to a different well-studied class of graphs. A graph $G$ is *perfect* if for every induced subgraph $H$ of $G$, the chromatic number (the smallest number of colours needed to colour vertices of $H$ such that adjacent vertices are coloured differently) of $H$ is equal to the size of the largest clique in $H$. The Strong Perfect Graph Theorem states that a graph $G$ is perfect if, and only if, $G$ does not contain any cycle of odd length $C_{2k+1}$, $k \geq 2$, (known as a hole) nor any complement of a cycle of odd length $\overline{C}_{2k+1}$, $k \geq 2$, (known as an antihole) as an induced subgraph [30]. It follows that a graph is perfect if, and only if, its complement is perfect. Figure 4 shows couples of forbidden induced subgraphs from the infinite class of forbidden induced subgraphs that characterises perfect graphs. It is known that MWIS in perfect graphs is solvable in polynomial time [31]. Moreover, perfect graphs can be recognised in polynomial time [32]. The combination of these results gives:

**Theorem 3.1.** *The class of VCSP instances with crisp binary and arbitrary unary constraints whose micro-structure complement is either perfect, fork-free or apple-free is tractable.*

The tractability of VCSPs with perfect micro-structure (complement) and soft unary constraints was also independently pointed out by Takhanov [33] and Jebara [34].

It is not possible to express all properties of VCSP or CSP instances uniquely in terms of properties of the micro-structure, since the micro-structure is a graph which does not contain the important information telling us which of its vertices correspond to assignments to the same variables. This information is, however, contained in the coloured micro-structure. This is why in the rest of the paper we consider properties of the coloured micro-structure in order to define novel tractable hybrid classes.

---

[2]Note that the class of apple-free graphs is the first class mentioned in this section for which it is not obvious that the *recognition* problem is solvable in polynomial time. However, the polynomial-time algorithm given in [29] is robust in the sense that it does not require the input graph $G$ to be apple-free; the algorithm either finds an independent set of maximum weight in $G$, or reports that $G$ is not apple-free.

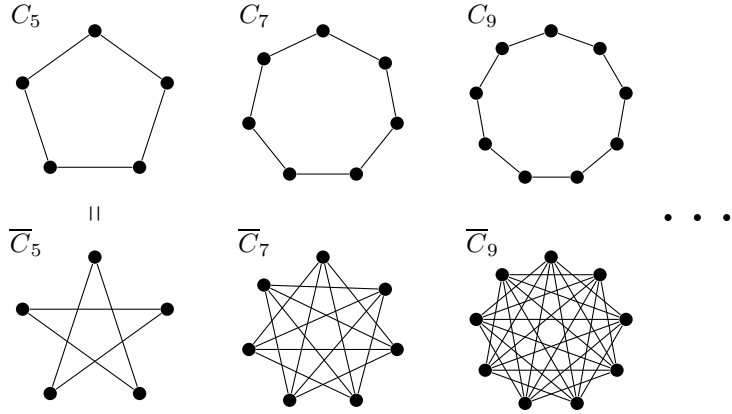

Figure 4: An infinite class of minimal non-perfect graphs.

# 4 Polynomially-bounded search

The tractable classes discussed in Section 3 can be defined by an objective function which is of the form $F_{fin} + F_{inf}$, where $F_{fin}$ is a finite-valued function (the sum of unary cost functions) and $F_{inf}$ is a $\{0, \infty\}$-valued function (whose micro-structure complement is perfect, fork-free or apple-free). In this section, we study other ways of defining tractable hybrid classes of VCSPs with an objective function of the form $F_{fin} + F_{inf}$, but where $F_{fin}$ is no longer simply the sum of unary cost functions.

## 4.1 Polynomial number of solutions

Notice that the crisp constraints given by $F_{inf}$ must define a tractable class of CSPs. If, moreover, the crisp constraints are so strong that the total number of feasible solutions (i.e. assignments of finite cost) is polynomially-bounded and these feasible solutions can be found in polynomial time, then the finite component $F_{fin}$ of the objective function can be arbitrary: an optimal assignment can be found in polynomial time by simply exhausting over all feasible solutions.

**Example 4.1.** Let $\mathcal{P}$ be a binary VCSP instance and let $M$ be the coloured micro-structure of the crisp component $F_{inf}$ of $\mathcal{P}$. Assuming that $M$ does not contain the pattern from Figure 5 (a) as an induced substructure, $\mathcal{P}$ has only polynomially many feasible solutions, as any assignment to two variables, say $i$ and $j$, can be extended to at most one assignment to another variable, say $k$. This can be clearly generalised to arbitrary fixed $k \geq 3$; the forbidden pattern would consist of a clique of size $k-1$ on distinct variables of the micro-structure and another $2(k-1)$ edges connecting the vertices of the clique to two different values of the $k$-the variable. The case $k = 4$ is depicted in Figure 5 (b): any assignment to three variables has at most one assignment to another variable. For a fixed $k$, this condition is easily recognisable in polynomial time.

**Example 4.2.** A binary CSP instance is *functional* if there exists a variable ordering $<$ such that, for all $j \in \{2, \ldots, n\}$, there is some constraint $c_{ij}$ with $i < j$ which is functional from $i$ to $j$, (i.e., in the notation of VCSPs, for each $a \in D_i$, there is at most one $b \in D_j$ such that $c_{ij}(a, b) < \infty$). In a CSP instance $I$, a *root set* is a subset $Q$ of the variables such that $I$ becomes functional after instantiation of all variables in $Q$. David [35] showed that it is possible to find a minimum-cardinality root set in polynomial time. Let $G_I$ be the directed graph with $n$ vertices and an arc from $i$ to $j$ if $I$ contains a functional constraint from $i$ to $j$. Then the size of the minimum-cardinality root set of $I$ is simply the number of source nodes in the directed graph of the strongly connected components of $G_I$ [35].

The set of VCSP instances whose crisp component $F_{inf}$ of the objective function is a CSP with a root set of size bounded by $O(\log n)$ represents a tractable hybrid class, independently of the soft constraints $F_{fin}$.

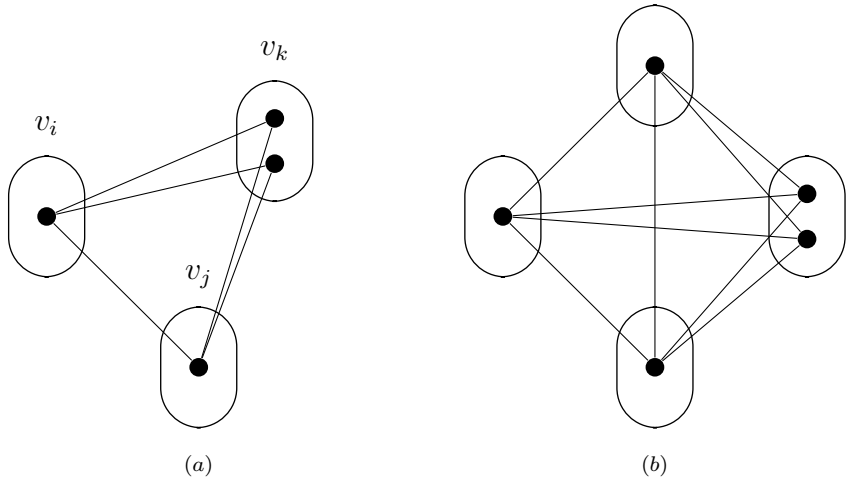

Figure 5: Forbidden patterns from Example 4.1.

## 4.2 Polynomial number of dead ends

We say that a set of VCSP instances is *conservative* if it is closed under arbitrary domain restrictions (i.e. the addition of unary crisp constraints). Another generic tractable case occurs when the CSP represented by $F_{inf}$ has a polynomial number of dead-ends in a backtracking search tree and the VCSP represented by $F_{fin}$ is a conservative tractable class. As we will now show, a polynomial number of dead-ends implies that the search space can be decomposed into a polynomial number of subspaces. A polynomial number of subspaces of solutions can be seen as a strict generalisation of a polynomial number of solutions (since each solution can be considered as a trivial subspace), but now we need to be able to minimise the finite component $F_{fin}$ in polynomial time in each of these subspaces of feasible solutions.

We introduce some notation. If $\mathcal{P}$ is a CSP instance and $j < k$ then $\mathcal{P}_{jk}^{cd}$ is the instance derived from $\mathcal{P}$ by

- keeping only the variables $1, \ldots, k$,
- reducing the domains of variables $v_j$ and $v_k$ to $\{c\}$ and $\{d\}$, respectively,
- discarding the binary constraints $c_{ik}$ for $i = j, \ldots, k-1$.

**Definition 4.3.** *A CSP instance $\mathcal{P}$ has the* polynomial number of dead-ends property *if for all $j < k \in \{1, \ldots, n\}$ and for all $c \in D_j$, $d \in D_k$, $c_{jk}(c, d) = \infty$ implies that the instance $\mathcal{P}_{jk}^{cd}$ has a polynomial number of solutions which can be found in polynomial time.*

**Proposition 4.4.** *Consider the class of VCSP instances whose objective function is of the form $F_{fin} + F_{inf}$, where $F_{fin}$ is a finite-valued function, $F_{inf}$ is a $\{0, \infty\}$-valued function, the VCSP instance corresponding to $F_{fin}$ is a conservative tractable class $\mathcal{C}$ and $F_{inf}$ corresponds to a binary CSP instance with the polynomial number of dead-ends property. Then this class is tractable.*

*Proof.* Let $\mathcal{P}$ be a CSP instance satisfying the polynomial number of dead-ends property. Consider the set $S_{jk}^{cd}$ of dead-ends in a backtracking search tree which are caused by the fact that the partial solution violates the constraint $c_{jk}$ due to the assignments $v_j = c$ and $v_k = d$. We identify a dead-end at level $k$ of the search tree with the corresponding assignment to variables $v_1, \ldots, v_k$. By the polynomial number of dead-ends property, it is possible to determine $S_{jk}^{cd}$ in polynomial time and, furthermore, $S_{jk}^{cd}$ is of polynomial size. Let $S$ represent the union of all the $S_{jk}^{cd}$ for all $j < k \in \{1, \ldots, n\}$ and all $c \in D_j$, $d \in D_k$. Each assignment $\langle a_1, \ldots, a_k \rangle$ in $S$ can be considered as an interval of non-solutions in the search space ordered by the standard lexicographic ordering (where each domain is ordered according to the order in which the backtracking algorithm exhausts over the domain). We can then calculate, in polynomial time, the union $U_S$ of these intervals (in the form of a set of intervals). The complement of $U_S$ (which can be returned as a set of intervals) is the set of solutions $sol(\mathcal{P})$ of $\mathcal{P}$.

To complete the proof, it suffices to show that the finite-valued objective function $F_{fin}$ can be minimised in polynomial time in each of the polynomial number of intervals in $sol(\mathcal{P})$ where $\mathcal{P}$ is the CSP equivalent to $F_{inf}$. Since $\mathcal{C}$ is a conservative tractable class, we only need to show that a lexicographic interval can be decomposed into the union of a polynomial number of sub-intervals each of which is defined by unary crisp constraints. Consider the interval $I$ consisting of all assignments between $\langle a_1, \ldots, a_n \rangle$ and $\langle b_1, \ldots, b_n \rangle$ in lexicographic order. Let $k \in \{0, \ldots, n\}$ be maximal such that $\forall i \in \{1, \ldots, k\}$, $a_i = b_i$. Then $I$ is the set of assignments with a prefix of one of the following forms: (1) $\langle a_1, \ldots, a_r, u \rangle$ with $u > a_{r+1}$ (for $r = k+1, \ldots, n-1$), (2) $\langle a_1, \ldots, a_k \rangle$ with $a_{k+1} < u < b_{k+1}$, (3) $\langle b_1, \ldots, b_r, u \rangle$ with $u < b_{r+1}$ (for $r = k+1, \ldots, n-1$). Each of these sub-intervals can be defined by crisp unary constraints. $\square$

**Example 4.5.** Consider the set of binary VCSP instances whose binary cost functions $c_{ij}$ are decomposable as the sum of a finite-valued cost function and a $\{0, \infty\}$-valued cost function $c_{ij}^{fin} + c_{ij}^{inf}$, where each $c_{ij}^{fin}$ is submodular and the crisp constraints $c_{ij}^{inf}$ satisfy the following rule: $\forall i < \max(j, k)$, $\forall a, b \in D_i$, $\forall c \in D_j$, $\forall d \in D_k$, $(a \neq b) \wedge (c_{ij}^{inf}(a, c) = c_{ij}^{inf}(b, c) = c_{ik}^{inf}(a, d) = c_{ik}^{inf}(b, d) = 0) \implies c_{jk}^{inf}(c, d) = 0$. This rule is equivalent to forbidding the pattern from Figure 6 as an induced substructure in the coloured micro-structure of $\mathcal{P}$, where $\mathcal{P}$ is the CSP defined by the crisp constraints $c_{ij}^{inf}$. (Note that the notion of an induced substructure requires that there is no edge between $(v_j, c)$ and $(v_k, d)$.) This rule implies that in $\mathcal{P}$, whenever $c_{jk}^{inf}(c, d) = \infty$, then $\mathcal{P}_{jk}^{cd}$ has at most one solution since in each domain $D_i$ ($i < \max(j, k)$) there is at most one value $a$ which is consistent with the two (variable,value) assignments $(v_j, c)$, $(v_k, d)$. This clearly implies the polynomial number of dead-ends property. Hence this set of VCSP instances is tractable by Proposition 4.4, since the class of VCSPs with submodular cost functions is conservative and tractable [14].

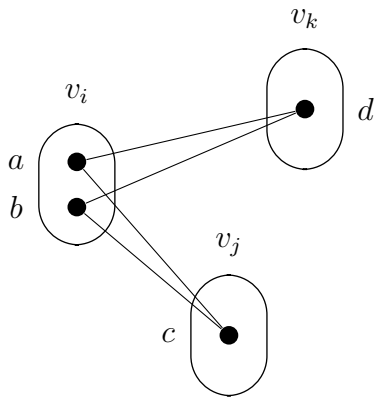

Figure 6: Forbidden pattern from Example 4.5.

### 4.3 Broken-triangle property

We have seen that forbidding certain patterns in the coloured micro-structure guarantees a polynomial number of solutions or dead ends, which in turn allows us to establish the tractability of certain classes of VCSPs. The recently discovered broken-triangle property (BTP) [11], which is also equivalent to forbidding a pattern in the coloured micro-structure of a CSP, guarantees that a feasible solution can be found in polynomial time. However, the BTP does not guarantee a polynomial number of solutions nor a polynomial number of dead ends. Indeed, in this section, we show that the hybrid tractable class of CSPs defined by the broken-triangle property is *not* extendible to VCSPs with soft unary constraints. A binary CSP instance $\mathcal{P}$ satisfies the broken-triangle property with respect to the variable ordering $<$ if, and only if, for all triples of variables $v_i, v_j, v_k$ such that $i < j < k$, if $c_{ij}(u,v) < \infty$, $c_{ik}(u,a) < \infty$ and $c_{jk}(v,b) < \infty$, then either $c_{ik}(u,b) < \infty$ or $c_{jk}(v,a) < \infty$. (In other words, every "broken" triangle $a - u - v - b$ can be closed.) This is equivalent to forbidding the pattern from Figure 7 as an induced substructure in the coloured micro-structure of $\mathcal{P}$.

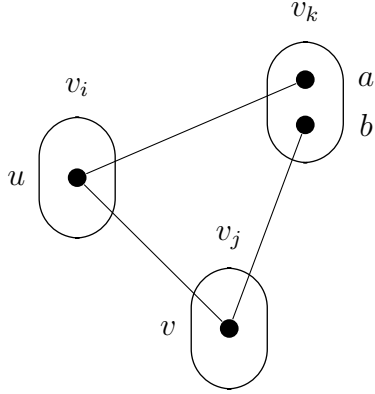

Figure 7: Forbidden pattern defining the broken-triangle property.

**Theorem 4.6.** *Assuming P$\neq$NP, the broken-triangle property cannot be extended to a tractable class including soft unary constraints.*

*Proof.* Let $\langle G, k \rangle$ be an instance of the decision version of the maximum independent set problem which consists in deciding whether there is an independent set of size at least $k$ in graph $G$. This problem is known to be NP-complete [36]. We now transform this instance into a binary VCSP instance with soft unary constraints that satisfies the broken-triangle property.

Every vertex of $G$ is represented by a Boolean variable $v_i$ where $D_i = \{0, 1\}$. We impose the constraint $\langle \langle v_i, v_j \rangle, \{\langle 0, 0 \rangle, \langle 0, 1 \rangle, \langle 1, 0 \rangle\} \rangle$ if the vertices corresponding to $v_i$ and $v_j$ are adjacent in $G$. Now the assignments satisfying all constraints are in one-to-one correspondence with independent sets $I$ in $G$, where vertex $i \in I$ if, and only if, $v_i = 1$. We also impose the soft unary constraints $\langle v_i, c_i \rangle$, where $c_i(x) = 1 - x$. The unary constraints ensure that the goal is to minimise the number of variables assigned value 0, which is the same as maximising the number of variables assigned value 1. Therefore, the constructed VCSP instance is equivalent to the given maximum independent set problem. It remains to show that the resulting VCSP instance satisfies the broken-triangle property with respect to some ordering. In fact, we show that it is satisfied with respect to any ordering. Take any three variables $v_i, v_j, v_k$. If either of the pairs of variables $\langle v_i, v_k \rangle$, $\langle v_j, v_k \rangle$ are not constrained, then the broken-triangle property is trivially satisfied. Assume therefore that these two constraints are present. The situation is illustrated in Figure 8. It can be easily checked that the broken-triangle property is indeed satisfied whether the constraint on $\langle v_i, v_j \rangle$ is $\{\langle 0, 0 \rangle, \langle 0, 1 \rangle, \langle 1, 0 \rangle\}$ (as shown in Figure 8) or the complete constraint. □

## 5 Joint-winner property

In this section we define the joint-winner property (see Figure 9), which is the key concept in this paper. We present several examples of known tractable problems that are generalised by the joint-winner property. We also study basic properties of VCSPs satisfying the joint-winner property as these will be important in designing a polynomial-time algorithm in Section 5.3.

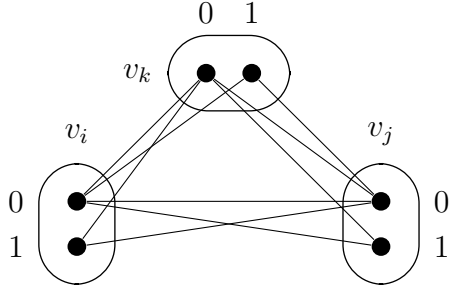

Figure 8: VCSP encoding maximum independent set.

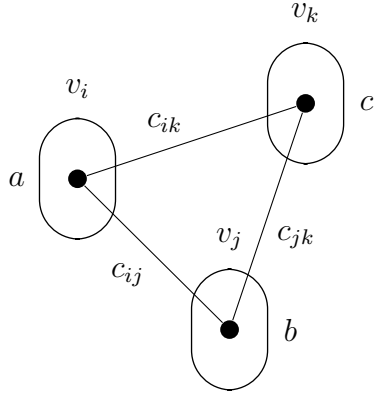

Figure 9: The joint-winner property: $c_{ij}(a,b) \geq \min(c_{ik}(a,c), c_{jk}(b,c))$.

**Definition 5.1** (Joint-winner property). *A triple of variables $\langle v_i, v_j, v_k \rangle$ satisfies the* joint-winner property *(JWP) if $c_{ij}(a,b) \geq \min(c_{ik}(a,c), c_{jk}(b,c))$ for all domain values $a \in D_i, b \in D_j, c \in D_k$.*

*A binary VCSP instance satisfies the* joint-winner property *if every triple of distinct variables of the instance satisfies the joint-winner property.*

Note that the joint-winner property places no restrictions on the unary constraints $c_i$ and is hence conservative.

### 5.1 Examples of the JWP

**Example 5.2.** The joint-winner property on a binary CSP instance amounts to forbidding the multiset of binary costs $\{\alpha, \infty, \infty\}$ (for $\alpha < \infty$) in any triangle formed by three assignments to distinct variables. Equivalently (given that the cost $\alpha$ corresponds to allowed assignments and thus non-edges in the micro-structure complement), a CSP instance $\mathcal{P}$ satisfies the joint winner property if, and only if, the coloured micro-structure complement of $\mathcal{P}$ forbids the structure in Figure 10 as an induced substructure. Since this combination can never occur on triples of variables $\langle v_i, v_j, v_k \rangle$ constrained by the three binary constraints $v_i \neq v_j \neq v_k \neq v_i$, the class of CSPs satisfying the joint-winner property generalises the ALLDIFFERENT constraint with arbitrary soft unary constraints. This generalisation is equivalent to allowing at most one assignment from each of a set of disjoint sets of (variable,value) assignments.

**Example 5.3.** Consider the (unweighted) MAX-2SAT problem with no repeated clauses, which is a well-known NP-complete problem [36, 37]. An instance $\phi$ of MAX-2SAT, where $\phi$ is a 2-CNF formula, can be seen as a Boolean binary $\{0,1\}$-valued VCSP instance. The joint-winner property is equivalent to the following condition on $\phi$: if $(l \vee l_1) \in \phi$ and $(l \vee l_2) \in \phi$, then $(l_1 \vee l_2) \in \phi$, where $l, l_1, l_2$ are three literals of distinct variables.

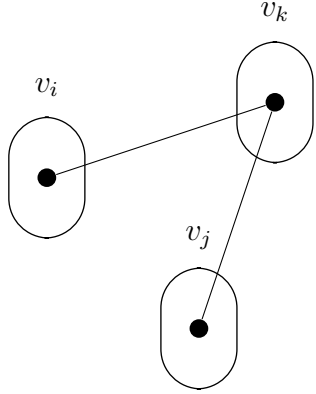

Figure 10: The crisp variant of the joint-winner property from Example 5.2.

As the next example shows, the well-known hybrid tractable class SOFTALLDIFF satisfies the JWP, thus showing that the JWP defines a *hybrid* tractable class (the tractability of the JWP being shown later in Section 5.3).

**Example 5.4.** One of the most commonly used global constraints is the ALLDIFFERENT constraint [38]. Petit et al. introduced a soft version of the ALLDIFFERENT constraint, SOFTALLDIFF [39]. They proposed two types of costs to be minimised: graph- and variable-based costs. The former counts the number of equalities, whilst the latter counts the number of variables violating an ALLDIFFERENT constraint. The algorithms for filtering these constraints, introduced in the same paper, were then improved by van Hoeve et al. [40].

It is easy to check that the graph-based variant of the SOFTALLDIFF constraint satisfies the joint-winner property. In this case for every $i$ and $j$, the cost function $c_{ij}$ is defined as $c_{ij}(a, b) = 1$ if $a = b$, and $c_{ij}(a, b) = 0$ otherwise. Take any three variables $v_i, v_j, v_k$ and $a \in D_i, b \in D_j, c \in D_k$. If $c_{ij}(a, b) = c_{jk}(b, c) = c_{ik}(a, c)$ (which means that that the domain values $a, b, c$ are all equal or all different), then the joint-winner property is satisfied trivially. If only one of of the costs is 1, then the joint-winner property is satisfied as well. Observe that due to the transitivity of equality it cannot happen that only one of the costs is 0.

In order to code the variable-based SOFTALLDIFF constraint as a binary VCSP $\mathcal{P}$, we can create $n$ variables $v'_i$ with domains $D_i \times \{1, 2\}$. The assignment $v'_i = (a, 1)$ means that $v_i$ is the first variable of the original CSP to be assigned the value $a$, whereas $v'_i = (a, 2)$ means that $v_i$ is assigned $a$ but it is not the first such variable. In $\mathcal{P}$ there is a crisp constraint which disallows $v'_i = v'_j = (a, 1)$ (for any value $a \in D_i \cap D_j$) for each pair of variables $i < j$ together with a soft unary constraint $c_i(a, k) = k - 1$ (for $k = 1, 2$) for each $i \in \{1, \ldots, n\}$. Hence at most one variable can be the first to be assigned $a$, and each assignment of the value $a$ to a variable (apart from the first) incurs a cost of 1. Again due to the transitivity of equality, it cannot happen that only one of the binary costs shown in the triangle of Figure 9 is zero, from which it follows immediately that the joint-winner property is satisfied in $\mathcal{P}$.

We remark that the class defined by the JWP is strictly bigger than SOFTALLDIFF, and hence our generic algorithm is not as efficient as tailor-made algorithms for SOFTALLDIFF.

**Example 5.5.** Suppose that $n$ jobs must be assigned to $d$ machines. Let $l_i(m)$ be the length of time required for machine $m$ to complete job $i$. If machine $m$ cannot perform job $i$, then $l_i(m) = \infty$. We use variable $v_i$ to represent the machine to which job $i$ is assigned. The set of jobs (which we denote by $S_m$) assigned to the same machine $m$ are performed in series in increasing order of their length $l_i(m)$. The aim is to minimise $T$ the sum, over all jobs, of their time until completion. If jobs $i$ and $j$ are assigned to the same machine, and $l_i(m) < l_j(m)$, then job $j$ will have to wait during the execution of job $i$, contributing a time of $l_i(m)$ to the sum $T$. This means that

$$T \;=\; \sum_{m=1}^{d} \Big( \sum_{i \in S_m} l_i(m) \;+\; \sum_{\substack{i,j \,\in\, S_m \\ i < j}} \min(l_i(m), l_j(m)) \Big).$$

In other words, optimal assignments of jobs to machines are given by solutions to the binary VCSP with unary constraints $c_i(m) = l_i(m)$ (representing the execution times of jobs) and binary constraints

$$c_{ij}(m,m') \;=\; \begin{cases} \min(l_i(m), l_j(m)) & \text{if } m = m' \\ 0 & \text{otherwise} \end{cases}$$

(representing the waiting times of jobs).

The joint-winner property $c_{ij}(a,b) \geq \min(c_{ik}(a,c), c_{jk}(b,c))$ is trivially satisfied in this VCSP instance if $a, b, c$ are not all equal, since in this case one of $c_{ik}(a,c), c_{jk}(b,c)$ is zero. It is also satisfied when $a = b = c = m$ since $\min(l_i(m), l_j(m)) \geq \min(\min(l_i(m), l_k(m)), \min(l_j(m), l_k(m)))$.

This problem has been shown solvable in polynomial time in [41, 42].

## 5.2 Structure of problems satisfying the JWP

The next lemma explains the reason for the name of the joint-winner property: in every triangle there is no unique smallest cost.

**Lemma 5.6.** *A VCSP instance satisfies the joint-winner property if, and only if, for all triples of distinct variables $\langle v_i, v_j, v_k \rangle$ and for all $a \in D_i$, $b \in D_j$, $c \in D_k$, two of the costs $c_{ij}(a,b), c_{ik}(a,c), c_{jk}(b,c)$ are equal and less than or equal to the third.*

*Proof.* Assume that the joint-winner property is satisfied on the triples of variables $\langle v_i, v_j, v_k \rangle$, $\langle v_j, v_k, v_i \rangle$ and $\langle v_k, v_i, v_j \rangle$, and write $\alpha = c_{ij}(a,b)$, $\beta = c_{ik}(a,c)$ and $\gamma = c_{jk}(b,c)$. Without loss of generality, let $\alpha = \min(\alpha, \beta, \gamma)$. From $\alpha \geq \min(\beta, \gamma)$, we can deduce that $\alpha = \min(\beta, \gamma)$ and hence that two of $\alpha, \beta, \gamma$ are equal and less than or equal to the third.

On the other hand, if two of $\alpha, \beta, \gamma$ are equal and less than or equal to the third, then $\min(\beta, \gamma) = \min(\alpha, \beta, \gamma) \leq \alpha$ and the JWP is satisfied. □

**Remark 5.7.** The joint-winner property is somewhat similar to the ultra-metric inequality. In this remark we will clarify the relationship.

An *ultra-metric* space is a set of points $M$ with an associated distance function (also called a metric) $d : M \times M \to \mathbb{R}$, where $\mathbb{R}$ is the set of real numbers, such that for $x, y, z \in M$ the following axioms hold:

1. $d(x,y) \geq 0$

2. $d(x,y) = 0$ if, and only if, $x = y$

3. $d(x,y) = d(y,x)$ (symmetry)

4. $d(x,z) \geq \min(d(x,y), d(y,z))$ (ultra-metric inequality).

It is well known that in an ultra-metric space, every triangle is isosceles (cf. Lemma 5.6).

Let $\mathcal{P}$ be a binary VCSP instance. Let $M = \{\langle v_i, a \rangle | a \in D_i\}$ be the set of vertices of the micro-structure of $\mathcal{P}$. Let $d(\langle v_i, a \rangle, \langle v_j, b \rangle) = c_{ij}(a,b)$ when $i \neq j$, and let $d(\langle v_i, a \rangle, \langle v_i, b \rangle) = \infty$ when $a \neq b$. Now if $\mathcal{P}$ satisfies the joint-winner property, then the ultra-metric inequality is satisfied by any three $x, y, z \in M$ if $x, y, z$ correspond to assignments to three different variables of $\mathcal{P}$. However, the JWP does not require, for instance, that $d(\langle v_i, a \rangle, \langle v_j, b \rangle) \geq \min(d(\langle v_i, a \rangle, \langle v_j, c \rangle), d(\langle v_j, b \rangle, \langle v_j, c \rangle))$. Hence $d$ does not satisfy the ultra-metric inequality on $M$.

We say that the binary constraint on variables $v_i, v_j$ has a *Z-configuration* on sub-domains $\{a,b\} \subseteq D_i$, $\{c,d\} \subseteq D_j$ if $c_{ij}(a,c), c_{ij}(b,c), c_{ij}(b,d) > c_{ij}(a,d)$. Since the JWP imposes a condition on exactly three variables, it imposes no condition on individual binary constraints. In particular, Z-configurations may occur. However, we will show that all Z-configurations can be eliminated by a simple pre-processing step. This will greatly simplify the structure of the instance allowing us to apply a min-cost max-flow algorithm to solve JWP instances.

We say that the pair of sub-domains $(S_i, S_j)$ (where $S_i \subseteq D_i$, $S_j \subseteq D_j$) is *independent of the other variables* if $\forall a,b \in S_i$, $\forall c,d \in S_j$, $\forall k \notin \{i,j\}$, $\forall e \in D_k$, $c_{ik}(a,e) = c_{ik}(b,e) = c_{jk}(c,e) = c_{jk}(d,e)$.

**Lemma 5.8.** *If a binary VCSP satisfying the JWP has a Z-configuration on sub-domains $\{a,b\} \subseteq D_i$, $\{c,d\} \subseteq D_j$, then $(\{a,b\},\{c,d\})$ is independent of the other variables.*

*Proof.* Without loss of generality, the presence of a Z-configuration means that $c_{ij}(a,c), c_{ij}(b,c), c_{ij}(b,d) > \beta$ where $c_{ij}(a,d) = \beta$. Consider some $k \notin \{i,j\}$ and some $e \in D_k$. By the JWP on the triple of assignments $\{\langle v_i, a \rangle, \langle v_j, d \rangle, \langle v_k, e \rangle\}$, we have either (1) $c_{ik}(a,e) \leq \beta$ or (2) $c_{jk}(d,e) \leq \beta$. In case (1), by repeated application of the JWP on the triples of assignments $\{\langle v_i, a \rangle, \langle v_j, c \rangle, \langle v_k, e \rangle\}$, $\{\langle v_i, b \rangle, \langle v_j, c \rangle, \langle v_k, e \rangle\}$, and $\{\langle v_i, b \rangle, \langle v_j, d \rangle, \langle v_k, e \rangle\}$, we can deduce that $c_{ik}(a,e) = c_{jk}(c,e) = c_{ik}(b,e) = c_{j,k}(d,e)$. In case (2), applying the JWP on the same triples, but in the reverse order, again allows us to deduce that $c_{ik}(a,e) = c_{jk}(c,e) = c_{ik}(b,e) = c_{jk}(d,e)$. □

**Lemma 5.9.** *In a VCSP instance satisfying the JWP, if $(S_i, S_j)$ (where $S_i \subseteq D_i$, $S_j \subseteq D_j$) is independent of the other variables, and $f \in D_i - S_i$ is such that $\exists c,d \in S_j$ such that $c_{ij}(f,c) \neq c_{ij}(f,d)$, then $(S_i \cup \{f\}, C_j)$ is independent of the other variables.*

*Proof.* Consider any variable $k \notin \{i,j\}$ and any $e \in D_k$. Since $(S_i, S_j)$ is independent of the other variables, $c_{jk}(c,e) = c_{jk}(d,e)$. By applying the JWP on the two triples of assignments $\{\langle v_i, f \rangle, \langle v_j, c \rangle, \langle v_k, e \rangle\}$ and $\{\langle v_i, f \rangle, \langle v_j, d \rangle, \langle v_k, e \rangle\}$, we can deduce that either $c_{ik}(f,e)$ is equal to $c_{jk}(c,e) = c_{jk}(d,e)$ or $c_{ik}(f,e)$ is equal to both of $c_{ij}(f,c)$ and $c_{ij}(f,d)$ (which is impossible since $c_{ij}(f,c) \neq c_{ij}(f,d)$). This shows that $(S_i \cup \{f\}, C_j)$ is independent of the other variables. □

We say that $(S_i, S_j)$ is *independent of other domain-values* if $\forall c,d \in S_j$, $\forall f \in D_i - S_i$, $c_{ij}(f,c) = c_{ij}(f,d)$ and $\forall a,b \in S_i$, $\forall g \in D_j - S_j$, $c_{ij}(a,g) = c_{ij}(b,dg)$.

Given a Z-configuration on $\{a,b\} \subseteq D_i$, $\{c,d\} \subseteq D_j$, we can use a simple algorithm to build a maximal pair of sub-domains $(S_i, S_j)$ which is independent of the other variables (and such that $\{a,b\} \subseteq S_i$, $\{c,d\} \subseteq S_j$). From Lemma 5.8, we can initialize $(S_i, S_j)$ to $(\{a,b\},\{c,d\})$. Then, by Lemma 5.9 we can simply keep adding $f \in D_i - S_i$ to $S_i$ if $\exists c',d' \in S_j$ such that $c_{ij}(f,c') \neq c_{ij}(f,d')$, and, by symmetry, we can keep adding $g \in D_j - S_j$ to $S_j$ if $\exists a',b' \in S_i$ such that $c_{ij}(a',g) \neq c_{ij}(b',g)$. When this process converges, $(S_i, S_j)$ is not only independent of the other variables but also independent of other domain-values.

Now let $p_0 \in S_i$ minimise $c_i(a)$, where $a \in S_i$. Similarly, let $q_0 \in S_j$ minimise $c_j(a)$, where $a \in S_j$. Furthermore, let $p_1 \in S_i$ and $q_1 \in S_j$ be the pair that minimises $c_i(a) + c_j(b) + c_{ij}(a,b)$ for $a \in S_i$ and $b \in S_j$. We replace the sub-domains $S_i, S_j$ by the single domain values $p, q$ where $c_i(p) = c_i(p_0)$, $c_j(q) = c_j(q_0)$, and $c_{ij}(p,q) = c_{ij}(p_1, q_1) + (c_i(p_1) - c_i(p_0)) + (c_j(q_1) - c_j(q_0))$ so that $c_i(p) + c_{ij}(p,q) + c_j(q) = c_{ij}(p_1, q_1) + c_i(p_1) + c_j(q_1)$. For all $f \in D_i - S_i$, we set $c_{ij}(f,q) = c_{ij}(f,q_0)$ and for all $g \in D_j - S_j$, we set $c_{ij}(p,g) = c_{ij}(p_0, g)$. Also $\forall k \notin \{i,j\}$, $\forall u \in D_k$, $c_{ik}(p,u) = c_{ik}(p_0, u) = c_{ik}(p_1, u)$ and $c_{jk}(q,u) = c_{jk}(q_0, u) = c_{jk}(q_1, u)$. The pair of domain values $p, q$ simulate either $p_0$ or $q_0$ if only one of the values $p, q$ is used but simulates $p_1, q_1$ if both values are used. We observe that this substitution preserves the JWP property. This is trivially true in all triangles involving only one of the assignments $p, q$, since the binary costs are identical to those for $p_0, q_0$. Since $S_i, S_j$ are independent of the other variables, any triangle of costs involving both assignments $p_1, q_1$ is isosceles. In particular, for any $u \in D_k$, where $k \neq i$ and $k \neq j$, $c_{ik}(p_1, u) = c_{jk}(q_1, u) = \alpha$ and $c_{ij}(p_1, q_1) \geq \alpha$. Since $c_{ik}(p,u) = c_{ik}(p_1, u) = c_{jk}(q_1, u) = c_{jk}(q,u)$ and $c_{ij}(p,q) \geq c_{ij}(p_1, q_1)$, the JWP property is preserved.

This substitution operation is guaranteed to preserve at least one optimal solution. This is because, by our choice of $p_0, q_0, p_1, q_1$, (1) in any assignment $x$ (to all $n$ variables) in which $x_i \in S_i$, $x_j \notin S_j$, replacing $x_i$ by $p_0$ (or by $p$ which is equivalent) cannot increase its cost, (2) in any assignment $x$ in which $x_i \notin S_i$, $x_j \in S_j$, replacing $x_j$ by $q_0$ (or by $q$) cannot increase its cost, and (3) in any assignment $x$ in which $x_i \in S_i$, $x_j \in S_j$, simultaneously replacing $x_i$ by $p_1$ and $x_j$ by $q_1$ (or replacing $x_i$ by $p$ and $x_j$ by $q$) cannot increase its cost.

We say that a VCSP instance satisfying the JWP is *Z-free* it contains no Z-configurations. We know that a VCSP satisfying the JWP can be made Z-free by applying the above pre-processing step to eliminate all Z-configurations. This pre-processing is clearly polynomial-time. A naive algorithm requires $O(n^2d^4)$ time to detect all Z-configurations. For a given pair $(S_i, S_j)$, when a new domain value $g$ is added to $S_j$ we have to test for all $f \in D_i - S_i$ whether $\exists c \in S_j$ such that $c_{ij}(f,c) \neq c_{ij}(f,g)$. This is an $O(d^2)$ operation. By a simple counting argument, we can see that the total number of times some domain value is added to some set $S_i$ is less than the total number of (variable,value) assignments which are eliminated by the substitution operation described above, which, in turn, is bounded above by $nd$. Thus the total complexity of the pre-processing operation is $O(n^2d^4 + nd^3)$ and hence $O(n^2d^4)$.

**Definition 5.10.** *Let $\mathcal{P}$ be a binary VCSP instance. A subgraph $C = \langle V_C, E_C \rangle$ of the micro-structure of $\mathcal{P}$ is an* assignment-clique *if $C$ forms a clique in the micro-structure of $\mathcal{P}$ when all edges between assignments to the same variable are added to $E_C$, i.e. $\langle V_C, E_C \cup S_C \rangle$ is a clique, where $S_C = \{\{\langle v_i, a \rangle, \langle v_i, b \rangle\} : \langle v_i, a \rangle, \langle v_i, b \rangle \in V_C$ and $a \neq b\}$.*

**Lemma 5.11.** *Let $\mathcal{P}$ be a Z-free binary VCSP instance satisfying the JWP. Then, for a fixed $\alpha$, the edges of the micro-structure of $\mathcal{P}$ corresponding to binary costs of at least $\alpha$, together with the corresponding vertices, form non-intersecting assignment-cliques.*

*Proof.* Since $\mathcal{P}$ is Z-free, within a single binary constraint any connected set of edges of the micro-structure of $\mathcal{P}$ corresponding to binary costs of at least $\alpha$ form a clique. For a contradiction let us assume that the edges of the micro-structure of $\mathcal{P}$ corresponding to binary costs of at least $\alpha$ do not form non-intersecting assignment-cliques. Hence there are three vertices $\langle v_i, a \rangle, \langle v_j, b \rangle, \langle v_k, c \rangle$ (where $i, j, k$ are distinct) of the micro-structure such that $c_{ij}(a,b) \geq \alpha$, $c_{ik}(a,c) \geq \alpha$, and $c_{jk}(b,c) < \alpha$. But this is in contradiction with Lemma 5.6. □

In the rest of the paper, when we refer to an assignment-clique $C_\alpha$ of edges in the micro-structure corresponding to binary costs of at least $\alpha$, we implicitly assume that $C_\alpha$ is maximal.

**Lemma 5.12.** *Let $\mathcal{P}$ be a Z-free binary VCSP instance satisfying the JWP. Let $C_\alpha$ be an assignment-clique in the micro-structure of $\mathcal{P}$ corresponding to binary costs of at least $\alpha$, and $C_\beta$ an assignment-clique in the micro-structure of $\mathcal{P}$ corresponding to binary costs of at least $\beta$. If $C_\alpha$ intersects $C_\beta$ and $\alpha \leq \beta$, then $C_\alpha \supseteq C_\beta$.*

*Proof.* Suppose that $C_\alpha$ and $C_\beta$ intersect and $\alpha \leq \beta$. If $\alpha = \beta$, the claim is satisfied trivially by Lemma 5.11, so we can suppose that $\alpha < \beta$. For a contradiction, assume that $C_\alpha \not\supseteq C_\beta$. By our assumptions, $\exists \langle v_i, a \rangle \in C_\alpha \cap C_\beta$ and $\exists \langle v_j, b \rangle \in C_\beta \setminus C_\alpha$. Since $C_\beta$ is an assignment-clique, we must have $c_{ij}(a,b) \geq \beta > \alpha$. Thus, by Lemma 5.11, the edge $\{\langle v_j, b \rangle, \langle v_i, a \rangle\}$ is part of an assignment-clique $C'_\alpha$ of edges of cost at least $\alpha$ (but not $C_\alpha$ since $\langle v_j, b \rangle \notin C_\alpha$). But then $C_\alpha$ and $C'_\alpha$ intersect at $\langle v_i, a \rangle$ which contradicts Lemma 5.11. □

## 5.3 Algorithm

In this section we present a polynomial-time algorithm for solving Z-free binary VCSPs satisfying the joint-winner property. The algorithm is an extension of a reduction to the standard max-flow/min-cut problem that has been used for flow-based soft global constraints [43, 40, 44].

First we review some basics on flows in graphs (see [45] for more details). Let $G = (V, A)$ be a directed graph with vertex set $V$ and arc set $A$. To each arc $a \in A$ we assign a *demand/capacity* function $[d(a), c(a)]$ and a *weight* function $w(a)$. Let $s, t \in V$. A function $f : A \to \mathbb{Q}$ is called an $s - t$ *flow* (or a flow) if

- $f(a) \geq 0$ for all $a \in A$;
- for all $v \in V \setminus \{s, t\}$, $\sum_{a=(u,v) \in A} f(a) = \sum_{a=(v,u) \in A} f(a)$ (flow conservation).

We say that a flow is *feasible* if $d(a) \leq f(a) \leq c(a)$ for each $a \in A$. We define the *value* of flow $f$ as $val(f) = \sum_{a=(s,v) \in A} f(a) - \sum_{a=(v,s) \in A} f(a)$. We define the *cost* of flow $f$ as $\sum_{a \in A} w(a)f(a)$. A *minimum-cost flow* is a feasible flow with minimum cost.

Algorithms for finding the minimum-cost flow of a given value are described in [46, Chapter 12] and [45]. Given a network $G$ with integer demand and capacity functions, the *successive shortest path algorithm* [46], can be used to find a feasible $s-t$ flow with value $\alpha$ and minimum cost in time $O(\alpha \cdot SP)$, where $SP$ is the time to compute a shortest directed path in $G$.

Given a Z-free binary VCSP $\mathcal{P}$ satisfying the JWP, we construct a directed graph $G_\mathcal{P}$ whose minimum-cost integral flows of value $n$ are in one-to-one correspondence with the solutions to $\mathcal{P}$. Apart from the source node $s$, $G_\mathcal{P}$ has three types of node:

1. a variable node $v_i$ ($i = 1, \ldots, n$) for each variable of $\mathcal{P}$;

2. an assignment node $\langle v_i, a \rangle$ ($a \in D_i$, $i = 1, \ldots, n$) for each possible variable-value assignment in $\mathcal{P}$;

3. an assignment-clique node $C_\alpha$ for each *maximal* assignment-clique of edges in the micro-structure of $\mathcal{P}$ corresponding to binary costs of at least $\alpha$. (The subscript $\alpha$ is equal to the minimum cost of edges in the assignment-clique and, where necessary, we use $C_\alpha$, $C'_\alpha$, ... to denote the distinct non-intersecting assignment-cliques corresponding to the same value of $\alpha$.)

In $G_\mathcal{P}$ there is an arc $(s, v_i)$ for each variable $v_i$ of $\mathcal{P}$. For all variables $v_i$ and for each $a \in D_i$, there is an arc $(v_i, \langle v_i, a \rangle)$ and also an arc from $\langle v_i, a \rangle$ to the assignment-clique $C_\alpha$ containing $\langle v_i, a \rangle$ such that $\alpha$ is maximal ($C_\alpha$ is unique by Lemma 5.12).

We say that assignment-clique $C_\beta$ is the *father* of assignment-clique $C_\alpha$ if it is the minimal assignment-clique which properly contains $C_\alpha$, i.e. $C_\alpha \subset C_\beta$ (and hence $\alpha > \beta$) and $\nexists C_\gamma$ such that $C_\alpha \subset C_\gamma \subset C_\beta$ ($C_\beta$ is unique by Lemma 5.12). In $G_\mathcal{P}$, for each assignment-clique $C_\alpha$ with father $C_\beta$, there is a bundle of arcs from $C_\alpha$ to $C_\beta$ consisting of $r$ arcs $e_i$ ($i = 1, \ldots, r$), where $r$ is the number of vertices in the assignment-clique $C_\alpha$. The weight of arc $e_i$ from $C_\alpha$ to $C_\beta$ is $w(e_i) = (i-1)(\alpha - \beta)$. (If $\alpha = \infty$ then there is a single arc of weight 0; the arcs of weight $\infty$ can simply be omitted.) We identify the sink node $t$ with the assignment-clique $C_0$ consisting of all edges in the micro-structure (since all binary costs are at least 0).

Each arc has demand 0 and capacity 1 except for arcs $(s, v_i)$ which have both demand 1 and capacity 1 (this forces a flow of exactly 1 through each variable node $v_i$). Weights of all arcs are 0 except for arcs going from an assignment-clique node to its father assignment-clique node, as described above, and arcs from a variable node $v_i$ to an assignment node $\langle v_i, a \rangle$ which have a weight of $c_i(a)$.

We show below that integral flows in the constructed network are in one-to-one correspondence with assignments in $\mathcal{P}$, but first we give an example.

**Example 5.13.** We show the general construction on a simple example. Let $\mathcal{P}$ be a VCSP instance with variables $v_1, v_2, v_3$ and $D_1 = D_2 = \{a, b\}$, $D_3 = \{a\}$. The coloured micro-structure of $\mathcal{P}$ is shown in Figure 11. Missing edges have cost 0. There are two assignment-cliques corresponding to cost at least 1: $C_1 = \{\langle v_1, a \rangle, \langle v_2, a \rangle, \langle v_3, a \rangle\}$ (in solid red in Figure 11) and $C'_1 = \{\langle v_1, b \rangle, \langle v_2, b \rangle\}$ (in dotted blue in Figure 11); and one assignment-clique corresponding to cost at least 2: $C_2 = \{\langle v_1, a \rangle, \langle v_2, a \rangle\}$ (in dashed green in Figure 11). The network corresponding to instance $\mathcal{P}$ is shown in Figure 12: demands and capacities are in square brackets, and weights of arcs without numbers are 0. The bold red edges represent flow $f$ corresponding to the assignment $v_1 = v_2 = v_3 = a$ with total cost 4, which is the same as the cost of $f$.

We now prove that integral flows $f$ in $G_\mathcal{P}$ are in one-to-one correspondence with assignments in the VCSP $\mathcal{P}$ and, furthermore, that the cost of $f$ is equal to the cost in $\mathcal{P}$ of the corresponding assignment.

All feasible flows have value $n$ since all $n$ arcs $(s, v_i)$ leaving the source have both demand and capacity equal to 1. Integral flows in $G_\mathcal{P}$ necessarily correspond to the assignment of a unique value $a_i$ to each variable $v_i$ since the flow of 1 through node $v_i$ must traverse a node $\langle v_i, a_i \rangle$ for some unique $a_i \in D_i$. It remains to show that for every assignment $\langle a_1, \ldots, a_n \rangle$ to $\langle v_1, \ldots, v_n \rangle$ which is feasible (i.e. whose cost in $\mathcal{P}$ is finite), there is a corresponding minimum-cost integral feasible flow $f$ in $G_\mathcal{P}$ of cost $\sum_i c_i(a_i) + \sum_{i<j} c_{ij}(a_i, a_j)$.

For each arc $a$ which is incoming to or outgoing from $\langle v_i, u \rangle$ in $G_\mathcal{P}$, let $f(a) = 1$ if $u = a_i$ and 0 otherwise. We denote the number of assignments $\langle v_i, a_i \rangle$ in assignment-clique $C_\alpha$ by $N(C_\alpha) = |\{\langle v_i, a_i \rangle \in C_\alpha : 1 \leq i \leq$ $n\}|$. By construction, each assignment-clique node $C_\alpha$ in $G_\mathcal{P}$ only has outgoing arcs to its father assignment-clique. For the outgoing arc $a$ of weight $i$ from $C_\alpha$ to its father assignment-clique, let $f(a) = 1$ if $N(C_\alpha) > i$ and 0 otherwise. This simply means that the outgoing arcs (each of capacity 1) from $C_\alpha$ are used in increasing order of their weight, one per assignment $\langle v_i, a_i \rangle \in C_\alpha$. This is clearly a minimum-cost flow corresponding to the assignment $\langle a_1, \ldots, a_n \rangle$.

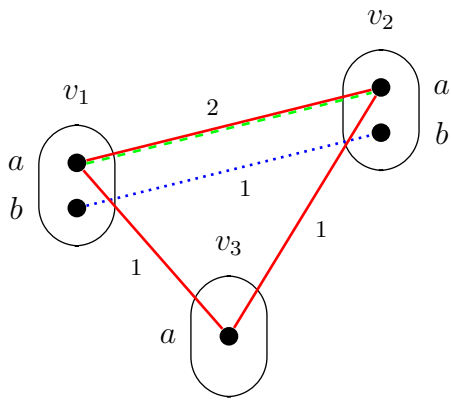

Figure 11: Micro-structure of $\mathcal{P}$ described in Example 5.13.

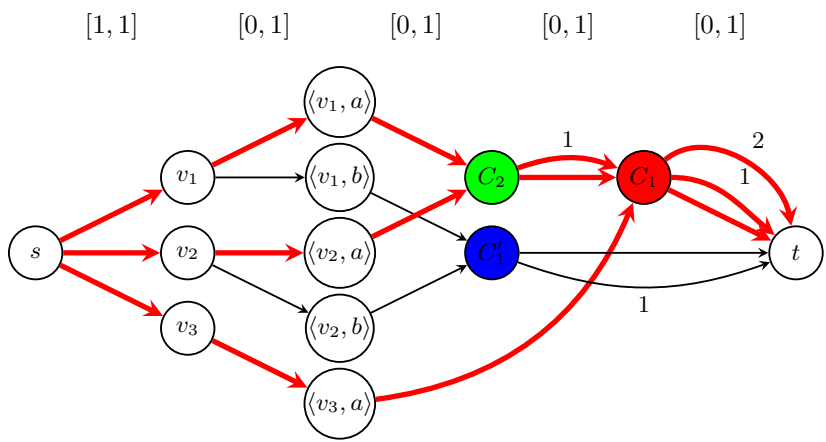

Figure 12: Network $G_\mathcal{P}$ corresponding to the VCSP $\mathcal{P}$ of Example 5.13.

Let $cf(C_\alpha)$ denote the cost $\beta$ of the father assignment-clique $C_\beta$ of $C_\alpha$. The cost of flow $f$ is given by

$$\sum_{i=1}^{n} c_i(a_i) + \sum_{C_\alpha} \sum_{i=1}^{N(C_\alpha)} (i-1)(\alpha - cf(C_\alpha))$$
$$= \sum_{i=1}^{n} c_i(a_i) + \sum_{C_\alpha} \sum_{i=1}^{N(C_\alpha)-1} i(\alpha - cf(C_\alpha))$$
$$= \sum_{i=1}^{n} c_i(a_i) + \sum_{C_\alpha} \frac{(N(C_\alpha)-1)N(C_\alpha)}{2}(\alpha - cf(C_\alpha)).$$

This corresponds precisely to the cost of the assignment $\langle a_1, \ldots, a_n \rangle$ in $\mathcal{P}$, since in an assignment-clique $C_\alpha$ with father assignment-clique $C_\beta$, each of the $(N(C_\alpha) - 1)N(C_\alpha)/2$ binary constraints contributes a cost of $\alpha - \beta$ over and above the cost of $\beta$ for each of the edges in $C_\beta$.

For the rest of this section, let $d = \max_{1 \le i \le n} |D_i|$ be the size of the largest domain. Given a VCSP instance satisfying the joint-winner property, there are clearly at most $|D_i| \times |D_j|$ different costs in the cost function $c_{ij}$. Hence in total there are at most $O(n^2 d^2)$ different maximal assignment-cliques $C_\alpha$. We now improve this upper bound to $O(nd)$.

**Lemma 5.14.** *A Z-free binary VCSP instance $\mathcal{P}$ satisfying the joint-winner property has at most $2nd - 1$ different maximal assignment-cliques $C_\alpha$.*

*Proof.* Consider binary VCSP instances $\mathcal{P}$ satisfying the joint-winner property and whose micro-structure has $N = nd$ vertices. Let $A_N$ be the maximum number of assignment-cliques $C_\alpha$ in such instances $\mathcal{P}$. We prove by induction that $A_N \le 2N - 1$. Clearly, $A_1 = 1$. Consider a micro-structure of size $N$. Consider any assignment-clique $C_\alpha$ of $\mathcal{P}$ which does not include all (variable,value) assignments of $\mathcal{P}$ but which is maximal in the sense that there is no other such assignment-clique $C_\beta \supset C_\alpha$. Let $r = |C_\alpha|$. By Lemma 5.12 and by maximality of $C_\alpha$, no other assignment-clique $C_\beta$ can intersect both $C_\alpha$ and its complement except an assignment-clique containing all the vertices of the micro-structure of $\mathcal{P}$. Hence we can partition the micro-structure of $\mathcal{P}$ (creating two VCSPs with micro-structures of size $r$ and $N - r$) and obtain the following recurrence: $A_N \le 1 + A_r + A_{N-r}$. From the inductive hypothesis, $A_r \le 2r - 1$ and $A_{N-r} \le 2(N - r) - 1$. Since $1 + 2r - 1 + 2(N - r) - 1 = 2N - 1$, it follows that $A_N \le 2N - 1$. □

We show that instances satisfying the joint-winner property can be recognised and solved in cubic time.

**Theorem 5.15.** *VCSPs satisfying the joint-winner property are solvable in polynomial time.*

*Proof.* From Definition 5.1, recognition can be achieved in $O(n^3 d^3)$ time. We have seen that a VCSP satisfying the JWP can be made Z-free in $O(n^2 d^4)$ time.

To solve a Z-free VCSP satisfying the JWP, we create a vertex for each of the assignment-cliques corresponding to binary costs of at least $\alpha$. By Lemma 5.14, there are at most $O(nd)$ different assignment-cliques. So our network has $O(nd + nd + n + 2) = O(nd)$ vertices. The network can be built in $O(n^3 d^3)$ time by $O(nd)$-times invoking depth-first search on the micro-structure of the instance, which is of size $O(n^2 d^2)$.

The result follows from the fact that a polynomial-time algorithm exists for determining a minimum-cost maximum flow in a network. In particular, using the successive shortest path algorithm, the running time is $O(n \cdot SP)$, where $SP$ is the time to compute a shortest directed path in the network [46, 45]. Using Fibonacci heaps, this is $O(n(n^2 d^2 + nd \log(nd))) = O(n^3 d^2)$. Hence, the complexity to solve a Z-free JWP instance is $O(n^3 d^3)$. The total complexity, including pre-processing is thus $O(n^2 d^3 (n + d))$. □

## 5.4 Maximality

Tractable classes defined by structural or language restrictions are often shown to be maximal. That is, any extension of the class is NP-hard. We consider that a hybrid tractable class defined by a set of possible combinations of costs within a subproblem is maximal if extending it to include any other single combination of costs renders the problem NP-hard. In particular, since JWP is defined on 3-variable subproblems, we call an instance *3-maximal* if extending it to include any other single combination of costs on 3 variables renders the problem NP-hard. In this section we show that the joint-winner property defines a 3-maximal tractable class.

We first consider the special case in which all costs belong to $\{\alpha, \beta\}$ (for some fixed distinct costs $\alpha < \beta$). In this case, the joint-winner property defines a maximal tractable class, except for the case when the domain size is 2 and all costs belong to $\{\alpha, \infty\}$, where $\alpha < \infty$. This latter case is just the well-known tractable class of Boolean binary CSPs [47].

**Theorem 5.16.** *If all costs belong to $\{\alpha, \beta\}$ (for some fixed distinct costs $\alpha < \beta$), then the joint-winner property defines a maximal tractable class provided $d > 2$ or $(d \geq 2) \wedge (\beta < \infty)$, where $d$ is the maximum domain size.*

*Proof.* To prove maximality we have to show the NP-hardness of the set of instances defined by the fact that in each triangle the triple of costs either satisfies the joint-winner property or is just one other fixed combination. Since all costs belong to $\{\alpha, \beta\}$ where $\alpha < \beta$, from Definition 5.1, the only situation forbidden by the JWP is that there are 3 variables $v_i, v_j, v_k$ and domain values $a \in D_i, b \in D_j, c \in D_k$ such that $c_{ij}(a, b) = \alpha$ and $c_{ik}(a, c) = c_{jk}(b, c) = \beta$. Hence extending the JWP means allowing all combinations of costs from $\{\alpha, \beta\}$ in all triangles.

If $\beta = \infty$, allowing all combinations of costs means that our instance allows all binary relations (corresponding to the set of finite-cost tuples) and hence we can encode any binary CSP. This is NP-complete if $d > 2$ [37].

If $\beta < \infty$, allowing all combinations of costs in $\{\alpha, \beta\}$ is equivalent to the set of instances of MAX-CSP in which no two constraints can have the same scope. The NP-hardness of this latter problem for $d \geq 2$ follows from the following reduction from MAX-CSP, which is a well-known NP-complete problem [36, 37]. A polynomial-time reduction of an instance $I$ of MAX-CSP into an equivalent instance $I'$ in which no two constraints have the same scope can be achieved by replacing each variable $v_i$ in $I$ by $M$ variables $v_i^j$ ($j = 1, \ldots, M$) in $I'$ constrained by a clique of equality constraints, where $M$ is greater than the total number of constraints in $I$. In the optimal solution to $I'$, variables $v_i^j$ ($j = 1, \ldots, M$) are necessarily assigned the same value (otherwise a cost of at least $M$ would be incurred). □

We now prove our main theorem concerning 3-maximality of the joint-winner property.

**Theorem 5.17.** *The joint-winner property defines a 3-maximal tractable class for any domain size $d \geq 2$.*

*Proof.* Let $\alpha < \beta \leq \gamma$ be any combination of costs which does not satisfy the JWP. To prove 3-maximality we have to show the NP-hardness of the set of instances defined by the fact that in each triangle the triple of costs either satisfies the joint-winner property or is $\{\alpha, \beta, \gamma\}$. We firstly consider the case $\gamma < \infty$.

It is well known that MAX-CSP is NP-hard [37]. Firstly, we show that binary MAX-CSP (coded as a binary VCSP with $\{0, 1\}$-valued cost functions) is NP-hard even on bipartite graphs. This follows from the fact that we can easily convert any instance of binary MAX-CSP into an instance of binary MAX-CSP on a bipartite graph, as follows. For each $c_{ij}(a, b) = 1$, introduce an extra Boolean variable $w_{ijab}$, and replace the cost of 1 for $(a, b)$ in $c_{ij}$ by costs of 1 for $(x_i, w_{ijab}) = (a, 0)$ and $(w_{ijab}, x_j) = (1, b)$ (all other costs being 0). In the resulting problem there is an obvious partition: (the original $x_i$ variables, the new $w_{ijab}$ variables).

Since $0 \leq \alpha < \beta < \infty$, any $\{0, 1\}$-valued VCSP instance is equivalent to the $\{\alpha, \beta\}$-valued instance obtained by applying the following operation to all cost functions: multiply by $\beta - \alpha$ and add $\alpha$. Now, given any instance of binary MAX-CSP on a bipartite graph (in the form of a $\{\alpha, \beta\}$-valued binary VCSP), we can add a constant cost function ($c_{ij}(x, y) = \gamma$ for all $x, y$) between any pair of variables within the same part without changing the problem. This just adds a constant cost to all solutions. In the resulting VCSP, all triangles of costs are $\{\gamma, \gamma, \gamma\}$, $\{\alpha, \alpha\gamma\}$, $\{\beta, \beta\gamma\}$ or $\{\alpha, \beta, \gamma\}$. Thus they either satisfy the JWP or are $\{\alpha, \beta, \gamma\}$. This polynomial reduction from binary MAX-CSP on bipartite graphs demonstrates NP-hardness in the case $\alpha < \beta < \infty$.

We now consider the case $\alpha < \beta < \gamma = \infty$. We demonstrate the NP-hardness of the set of VCSP instances such that in each triangle the triple of costs either satisfies the JWP or is $\{\alpha, \beta, \infty\}$. A VCSP remains invariant under addition of a constant finite cost and under scaling by a non-zero finite constant factor. Thus, for $\alpha < \beta < \infty$, a VCSP such that in each triangle the triple of costs either satisfies the JWP or is $\{\alpha, \beta, \infty\}$ is equivalent to a VCSP such that in each triangle the triple of costs either satisfies the JWP or is $\{0, 1, \infty\}$. We demonstrate NP-hardness of such VCSPs by the following polynomial-time reduction from MAX-2SAT. Let $I$ be an instance of MAX-2SAT with variables $v_i$ ($i = 1, \ldots, n$) and let $m_i$ be the number of clauses in which variable $v_i$ occurs. We construct a VCSP instance $\mathcal{P}$ containing $m_i$ copies $v_i^j$ ($j = 0, \ldots, m_i - 1$) of each variable $v_i$. For each $i = 1, \ldots, n$, the $m_i$ variables $v_i^j$ ($j = 0, \ldots, m_i - 1$) are joined together by a cycle of $\{0, \infty\}$-valued crisp constraints coding the inequalities $v_i^j \leq v_i^{j+1}$ (where the addition in the superscript is modulo $m_i$). Since these crisp constraints form a cycle, they are equivalent to $v_i^0 = \ldots = v_i^{m_i-1}$. It is easy to verify that this cycle of crisp constraints satisfies the JWP since no triangle of binary costs contains two infinite costs.

The 2SAT clauses of $I$ are replaced by corresponding $\{0, 1\}$-valued constraints in $\mathcal{P}$ such that the $j$th occurrence of $v_i$ in $I$ is replaced by $v_i^j$ in $\mathcal{P}$. By this construction, every triangle of variables in $\mathcal{P}$ involves at most one $\{0, 1\}$-valued constraint corresponding to a clause of $I$. It follows that the triple of costs $\{0, 1, 1\}$ does not occur in any triangle in $\mathcal{P}$. Since infinite cost can only occur in the non-intersecting cycles of crisp constraints, the triples of costs $\{0, \infty, \infty\}$ and $\{1, \infty, \infty\}$ also do not occur in any triangle in $\mathcal{P}$. Thus, in each triangle in $\mathcal{P}$ the triple of costs either satisfies the JWP or is $\{0, 1, \infty\}$. Since $\mathcal{P}$ is equivalent to $I$, and the reduction is clearly polynomial, this completes the proof of NP-hardness.

It remains to consider the case $\alpha < \beta = \gamma = \infty$. Analogously to the case $\alpha < \beta < \gamma = \infty$, it suffices to give a polynomial-time reduction from MAX-2SAT to the set of VCSP instances such that the costs in each triangle either satisfy the JWP or are $\{0, \infty, \infty\}$. Let $I$ be an instance of MAX-2SAT. We now build a VCSP instance $\mathcal{P}$ from $I$ by adding two new variables $x_{ij}, y_{ij}$ for each clause $l_1 \vee l_2$ of $I$ (where $l_1, l_2$ are literals). The clause $l_1 \vee l_2$ of $I$ is replaced in $\mathcal{P}$ by a $\{0, 1\}$-valued constraint on variables $x_{ij}, y_{ij}$ encoding the clause $\neg x_{ij} \vee \neg y_{ij}$ together with two crisp (i.e. $\{0, \infty\}$-valued constraints) encoding the clauses $l_1 \vee x_{ij}$ and $y_{ij} \vee l_2$. It is easy to verify that $\mathcal{P}$ is equivalent to $I$ and that the costs in each triangle in $\mathcal{P}$ either satisfy the JWP or are $\{0, \infty, \infty\}$. This polynomial reduction from MAX-2SAT completes the proof of NP-hardness for the final case $\alpha < \beta = \gamma = \infty$. □

## 5.5 Non-overlapping convexity property

Having studied in detail the joint-winner property on triangles of binary costs, a natural question is whether it can be generalised to more than three variables. By analysing our algorithm for Z-free binary VCSPs satisfying the joint-winner property (Section 5.3), we can extend the class of problems solvable in polynomial time using the same approach. This generalisation is no longer restricted to binary VCSPs. In a VCSP with constraints of arbitrary arity, the objective function to be minimised is the sum of cost functions whose arguments are subsets of arbitrary size of the variables $v_1, \ldots, v_n$. For notational convenience, in this section we interpret a solution $x$ (i.e. an assignment to the variables $v_1, \ldots, v_n$) as the set of (variable,value) assignments $\{(v_i, x_i) : i = 1, \ldots, n\}$.

For the following definition, we require the notion of a non-decreasing derivative of a discrete function. The derivative of the function $f : \{0, \ldots, s\} \to \overline{\mathbb{Q}_+}$ is *non-decreasing* if $f(m+2) - f(m+1) \geq f(m+1) - f(m)$ for all $m \in \{0, \ldots, s-2\}$ (where subtraction is extended to $\overline{\mathbb{Q}_+}$ by defining $\infty - \alpha = \infty$ for all $\alpha \in \overline{\mathbb{Q}_+}$). Two set $S, T$ are said to be *non-overlapping* if they are either disjoint or one is a subset of the other (i.e. $S \cap T = \emptyset$, $S \subseteq T$ or $T \subseteq S$).

**Definition 5.18.** *Let $\mathcal{P}$ be a VCSP instance. Let $C_1, \ldots, C_r$ be sets of (variable,value) assignments of $\mathcal{P}$ such that for all $i, j$, $C_i$ and $C_j$ are non-overlapping. Instance $\mathcal{P}$ satisfies the* non-overlapping convexity *property if the objective function of $\mathcal{P}$ can be written as $f_1(N(x, C_1)) + \ldots + f_r(N(x, C_r))$ such that each* $f_i : \{0, \ldots, s_i\} \to \overline{\mathbb{Q}}_+$ $(i = 1, \ldots, r)$ is a non-negative non-decreasing function with non-decreasing derivative, where $N(x, C_i) = |x \cap C_i|$ is the number of (variable,value) assignments in the solution $x$ which lie in $C_i$ and $s_i$ is the number of distinct variables occurring in the set of (variable,value) assignments $C_i$.

Observe that this definition allows any unary valued constraints, since for each (variable,value) assignment $(v_j, a)$ we can add the singleton $C_i = \{(v_j, a)\}$ which is necessarily either disjoint or a subset of any other set $C_k$ (and furthermore the corresponding function $f_i : \{0, 1\} \to \overline{\mathbb{Q}}_+$ is necessarily a non-negative function with a trivially non-decreasing derivative).

**Remark 5.19.** To express a binary Z-free VCSP satisfying the JWP in the more general form given in Definition 5.18, we add a set of (variable,value) assignments $C_i$ corresponding to each assignment-clique $C_\alpha$ in the micro-structure, with $f_i(m) = \binom{m}{2}(\alpha - \beta)$ for $m \geq 0$, where $C_\beta$ is the father assignment-clique of $C_\alpha$. It is easy to check that $f_i$ is non-negative, non-decreasing and has a non-decreasing derivative.

**Theorem 5.20.** *Any VCSP instance $\mathcal{P}$ satisfying the non-overlapping convex property can be solved in polynomial time.*

*Proof.* We can consider each set of (variable,value) assignments $C_i$ as an assignment-clique: since $C_i$ is a subset of the vertices of the micro-structure of $\mathcal{P}$, we just need to add all edges between those vertices $(v_j, a), (v_k, b)$ of $C_i$ representing assignments to distinct variables ($j \neq k$).

We build the same network as described in Section 5.3, but have to adjust the weights on edges between the assignment-cliques $C_i$. Recall that $s_i$ represents the number of distinct variables in $C_i$. Without loss of generality, we can assume that $f_i(0) = 0$. (If this is not the case, we can replace $f_i$ by $f'_i$ where $f'_i(m) = f_i(m) - f_i(0)$ to produce an equivalent problem with $f'_i(0) = 0$ and in which $f'_i$ is non-negative, non-decreasing and with a non-decreasing derivative.) Let $\Delta_i(m) = f_i(m) - f_i(m-1)$ for $m = 1, \ldots, s_i$. Since the $f_i$ ($1 \leq i \leq r$) are non-decreasing functions with a non-decreasing derivative, $\Delta_i$ is non-negative and non-decreasing. In the network, there are $s_i$ arcs $e_i^k$ ($k = 1, \ldots, s_i$) from $C_i$ to its (unique) father assignment-clique. The weight (cost) of arc $e_i^k$ is $\Delta_i(k)$.

Similarly to the construction in Section 5.3, arcs of weight $\infty$ can be omitted. The same argument and calculation as in Section 5.3 proves that the algorithm is correct. In particular, we need to show that given any feasible assignment $x$ (that is, of finite cost) to a binary VCSP instance $\mathcal{P}$ that satisfies the non-overlapping convex property, there is a corresponding minimum-cost integral feasible flow $f$ in $G_\mathcal{P}$ of cost $f_1(N(x, C_1)) + \ldots + f_r(N(x, C_r))$. The same construction of $f$ as in Section 5.3 gives us a flow $f$ of the following cost:

$$\sum_i \sum_{m=1}^{N(C_i)} \Delta_i(m) = \sum_i \big[ f_i(1) + (f_i(2) - f_i(1)) + \ldots$$
$$+ (f_i(N(C_i) - 1) - f_i(N(C_i) - 2))$$
$$+ (f_i(N(C_i)) - f_i(N(C_i) - 1)) \big]$$
$$= \sum_i f_i(N(C_i)).$$

which corresponds precisely to the cost $f_1(N(x, C_1)) + \ldots + f_r(N(x, C_r))$ of the assignment $x$. To see that this flow is a of minimum cost, among flows corresponding to the assignment $x$, observe that the only choices to be made concern the arcs $e_i^k$. Since the weights $\Delta_i(k)$ of these arcs are non-decreasing, a flow of value $v$ which takes the first $v$ arcs $e_i^k$ ($k = 1, \ldots, v$) is necessarily of minimum cost. $\square$

**Example 5.21.** An example of the non-overlapping convex property is the set of non-binary (i.e. arbitrary but fixed maximum arity) CSP (or MAX-CSP) instances with no overlapping nogoods, which we will now describe. A CSP is traditionally defined by a set of constraints each of which is given in the form of a scope and a relation containing the set of possible assignments to the variables in the scope of the constraint. Another way of specifying a CSP instance is by listing its nogoods, where a nogood is a set of (variable,value) assignments which cannot simultaneously be made. The nogoods are easily obtained from the complement of each constraint relation. A CSP instance with no overlapping nogoods is such that for all nogoods $N_i, N_j$, either $N_i \cap N_j = \emptyset$, $N_i \subset N_j$ or $N_j \subset N_i$.

To see that any CSP (or MAX-CSP) instance $I$ with non-overlapping nogoods $N_1, \ldots, N_r$ satisfies the non-overlapping convex property, for $i = 1, \ldots, r$, define $f_i$ by $f_i(|N_i|) = \infty$ (or $f_i(|N_i|) = 1$ in the case of MAX-CSP) and $f_i(m) = 0$ for $m < |N_i|$. Clearly each $f_i$ is a non-negative (non-decreasing) function with a non-decreasing derivative and the CSP (or MAX-CSP) instance $I$ is equivalent to the VCSP instance with objective function $f_1(|x \cap N_1|) + \ldots + f_r(|x \cap N_r|)$.

**Example 5.22.** An example of an optimisation problem satisfying the non-overlapping convex property is the following office assignment problem. Each of $n$ staff members, represented by $n$ variables, must be assigned an office. There are $m$ offices. At most $u_j$ people can be assigned office $j$. Unary cost functions can be used to express personal preferences of each staff member for each office. There are also non-overlapping groups of people $G_1, \ldots, G_g$ who we would prefer to assign to different offices (such as married couples, for example).

For each office there is a set of (variable,value) assignments $C_j$ consisting of all possible assignments of value $j$ to any variable. For each group $G_i$ and for each office $j$, there is a set of assignments $C_{ij} \subseteq C_j$ of members of this group to office $j$. Clearly, the sets of assignments $C_j$ ($j \in \{1, \ldots, m\}$) and $C_{ij}$ ($i \in \{1, \ldots, g\}$, $j \in \{1, \ldots, m\}$) are all non-overlapping. The function $f_j$ which imposes the capacity constraint for office $j$, given by $f_j(k) = \infty$ if $k > u_j$ (and $f_j(k) = 0$ otherwise), is non-negative, non-decreasing and has a non-decreasing derivative. A cost function $f_{ij}$ such as $f_{ij}(t) = t - 1$ (which is non-negative, non-decreasing and has a non-decreasing derivative) can be used to code the fact that we prefer that staff members from group $G_i$ do not share the same office.

**Example 5.23.** A commonly occurring problem in academia is the allocation of courses to teachers subject to timetabling constraints and the personal preferences of the teachers along with a criterion to avoid allocating too many hours to any of the teachers. In a simple version of this problem, the weekly timetable has been divided up into non-intersecting time-slots, each course involves giving only one lecture a week and one-hour time-slots have already been allocated to each course. The courses, numbered from 1 to $n$, correspond to the variables of the problem and must be assigned one of the $m$ teachers.

For each teacher $j \in \{1, \ldots, m\}$, let $C_j$ represent all assignments of teacher $j$ to any course. For each time-slot $s \in \{1, \ldots, t\}$ and each teacher $j \in \{1, \ldots, m\}$, let $C_{js}$ represent the set of assignments of teacher $j$ to courses allocated the time-slot $s$. Clearly, the sets of assignments $C_j$ ($j \in \{1, \ldots, m\}$) and $C_{js}$ ($j \in \{1, \ldots, m\}$ and $s \in \{1, \ldots, t\}$) are all non-overlapping. For each teacher $j$, the function $f_j$ corresponding to the set of assignments $C_j$ and given by $f_j(k) = \max(0, p_j(k - u_j))$ (where $u_j$ is the maximum number of courses teacher $j$ can give before they have to be paid overtime and $p_j$ is the amount teacher $j$ earns per hour of overtime) represents the total cost in overtime payments of teacher $j$. The function $f_{js}$, which codes the incompatibility of allocating teacher $j$ to two different courses during the same time-slot $s$, is given by $f_{sj}(0) = f_{sj}(1) = 0$ and $f_{js}(k) = \infty$ if $k > 1$. The functions $f_j$ and $f_{ij}$ are non-negative, non-decreasing and have a non-decreasing derivative. Thus this simple course-allocation problem can be expressed as a VCSP with the non-overlapping convex property.

# 6 Conclusions

We have studied hybrid tractability of soft CSPs. In particular, we have studied the tractability of sets of instances defined by properties of subproblems of size $k$. For $k = 2$, such properties can only define language classes. We have shown in this paper that it is possible to define a non-trivial tractable hybrid class by a property on subproblems of size $k = 3$. We have studied the tractable class of VCSPs defined by this property, known as the joint-winner property (JWP), as a necessary first step towards a general theory of tractability of optimisation problems which will eventually cover structural, language and hybrid reasons for tractability. Moreover, we have presented several other novel hybrid tractable classes of VCSPs.

The JWP is interesting in its own right since it is a proper extension to known tractable classes (such as VCSPs consisting of arbitrary unary constraints and non-intersecting SOFTALLDIFF constraints of arbitrary arity, as well as a machine scheduling problem).

We have demonstrated 3-maximality of the tractable class defined by the JWP. However, the existence of a larger tractable class subsuming JWP and defined by a rule on $k$-variable subproblems (for $k > 3$) is an interesting open question. Indeed, we have introduced a generalisation of the JWP which is solved by a similar algorithm but which allows soft constraints of arbitrary arity.

The most interesting open question brought out by this work is whether other tractable hybrid classes can be defined by properties of $k$-variable subproblems for $k \geq 3$.

# References

[1] A. Bulatov, A. Krokhin, P. Jeavons, Classifying the Complexity of Constraints using Finite Algebras, SIAM Journal on Computing 34 (3) (2005) 720–742. `doi:10.1137/S0097539700376676`. 1

[2] T. Feder, M. Vardi, The Computational Structure of Monotone Monadic SNP and Constraint Satisfaction: A Study through Datalog and Group Theory, SIAM Journal on Computing 28 (1) (1998) 57–104. `doi:10.1137/S0097539794266766`. 1

[3] R. Dechter, J. Pearl, Network-based Heuristics for Constraint Satisfaction Problems, Artificial Intelligence 34 (1) (1988) 1–38. `doi:10.1016/0004-3702(87)90002-6`. 1

[4] P. Jeavons, On the Algebraic Structure of Combinatorial Problems, Theoretical Computer Science 200 (1-2) (1998) 185–204. `doi:10.1016/S0304-3975(97)00230-2`. 1

[5] V. Dalmau, P. G. Kolaitis, M. Y. Vardi, Constraint Satisfaction, Bounded Treewidth, and Finite-Variable Logics, in: Proceedings of the 8th International Conference on Principles and Practice of Contraint Programming (CP'02), Vol. 2470 of Lecture Notes in Computer Science, Springer, 2002, pp. 310–326. `doi:10.1007/3-540-46135-3_21`. 1

[6] M. Grohe, The complexity of homomorphism and constraint satisfaction problems seen from the other side, Journal of the ACM 54 (1). `doi:10.1145/1206035.1206036`. 1

[7] R. Dechter, Constraint Processing, Morgan Kaufmann, 2003. 2

[8] D. A. Cohen, A New Class of Binary CSPs for which Arc-Constistency Is a Decision Procedure, in: Proceedings of the 9th International Conference on Principles and Practice of Constraint Programming (CP'03), Vol. 2833 of Lecture Notes in Computer Science, Springer, 2003, pp. 807–811. `doi:10.1007/b13743`. 2, 3

[9] D. Cohen, P. Jeavons, The complexity of constraint languages, in: F. Rossi, P. van Beek, T. Walsh (Eds.), The Handbook of Constraint Programming, Elsevier, 2006. 2

[10] T. K. S. Kumar, A framework for hybrid tractability results in boolean weighted constraint satisfaction problems, in: Proceedings of the 14th International Conference on Principles and Practice of Constraint Programming (CP'08), Vol. 5202 of Lecture Notes in Computer Science, Springer, 2008, pp. 282–297. `doi:10.1007/978-3-540-85958-1_19`. 2

[11] M. C. Cooper, P. G. Jeavons, A. Z. Salamon, Generalizing constraint satisfaction on trees: hybrid tractability and variable elimination, Artificial Intelligence 174 (9–10) (2010) 570–584. `doi:10.1016/j.artint.2010.03.002`. 2, 9

[12] T. Schiex, H. Fargier, G. Verfaillie, Valued Constraint Satisfaction Problems: Hard and Easy Problems, in: Proceedings of the 14th International Joint Conference on Artificial Intelligence (IJCAI'95), 1995. Available from: `http://dli.iiit.ac.in/ijcai/IJCAI-95-VOL1/pdf/083.pdf`. 2, 3

[13] S. Bistarelli, U. Montanari, F. Rossi, Semiring-based Constraint Satisfaction and Optimisation, Journal of the ACM 44 (2) (1997) 201–236. doi:10.1145/256303.256306. 2

[14] D. A. Cohen, M. C. Cooper, P. G. Jeavons, A. A. Krokhin, The Complexity of Soft Constraint Satisfaction, Artificial Intelligence 170 (11) (2006) 983–1016. doi:10.1016/j.artint.2006.04.002. 2, 9

[15] D. A. Cohen, M. C. Cooper, P. G. Jeavons, Generalising submodularity and Horn clauses: Tractable optimization problems defined by tournament pair multimorphisms, Theoretical Computer Science 401 (1-3) (2008) 36–51. doi:10.1016/j.tcs.2008.03.015. 2

[16] M. C. Cooper, S. Živný, A new hybrid tractable class of soft constraint problems, in: Proceedings of the 16th International Conference on Principles and Practice of Constraint Programming (CP'10), Vol. 6308 of Lecture Notes in Computer Science, Springer, 2010, pp. 152–166. 2

[17] S. L. Lauritzen, Graphical Models, Oxford University Press, 1996. 3

[18] M. J. Wainwright, M. I. Jordan, Graphical models, exponential families, and variational inference, Foundations and Trends in Machine Learning 1 (1-2) (2008) 1–305. doi:10.1561/2200000001. 3

[19] P. Jégou, Decomposition of Domains Based on the Micro-Structure of Finite Constraint-Satisfaction Problems, in: Proceedings of the 11th National Conference on Artificial Intelligence (AAAI'93), 1993, pp. 731–736. Available from: http://www.aaai.org/Papers/AAAI/1993/AAAI93-109.pdf. 3

[20] A. Z. Salamon, P. G. Jeavons, Perfect Constraints Are Tractable, in: Proceedings of the 14th International Conference on Principles and Practice of Constraint Programming (CP'08), Vol. 5202 of Lecture Notes in Computer Science, Springer, 2008, pp. 524–528. doi:10.1007/978-3-540-85958-1_35. 3

[21] L. W. Beineke, Characterizations of derived graphs, Journal of Combinatorial Theory 9 (2) (1970) 129–135. doi:10.1016/S0021-9800(70)80019-9. 4

[22] J. Edmonds, Paths, trees, and flowers, Canadian Journal of Mathematics 17 (1965) 449–467. Available from: http://www.math.ca./cjm/v17/cjm1965v17.0449-0467.pdf. 4

[23] J. Edmonds, Maximum Matching and a Polyhedron with 0, 1 Vertices, Journal of Research National Bureau of Standards 69 B (1965) 125–130. Available from: http://drop.io/jres69b. 4

[24] G. J. Minty, On maximal independent sets of vertices in claw-free graphs, J. Comb. Theory, Ser. B 28 (3) (1980) 284–304. doi:10.1016/0095-8956(80)90074-X. 4

[25] D. Nakamura, A. Tamura, A revision of Minty's algorithm for finding a maximum weighted stable set of a claw-free graph, Journal of the Operations Research Society of Japan 44 (2) (2001) 194–204. Available from: http://ci.nii.ac.jp/naid/110001183942/en. 4

[26] D. G. Corneil, H. Lerchs, L. Stewart-Burlingham, Complement reducible graphs, Discrete Applied Mathematics 3 (3) (1981) 163–174. doi:10.1016/0166-218X(81)90013-5. 6

[27] V. V. Lozin, M. Milanič, A polynomial algorithm to find an independent set of maximum weight in a fork-free graph, Journal of Discrete Algorithms 6 (4) (2008) 595–604. doi:10.1016/j.jda.2008.04.001. 6

[28] V. E. Alekseev, Polynomial algorithm for finding the largest independent sets in graphs without forks, Discrete Applied Mathematics 135 (1-3) (2004) 3–16. doi:10.1016/S0166-218X(02)00290-1. 6

[29] A. Brandstädt, V. V. Lozin, R. Mosca, Independent Sets of Maximum Weight in Apple-Free Graphs, SIAM Journal on Discrete Mathematics 24 (1) (2010) 239–254. doi:10.1137/090750822. 6

[30] M. Chudnovsky, N. Robertson, P. Seymour, R. Thomas, The strong perfect graph theorem, Annals of Mathematics 164 (1) (2006) 51–229. doi:10.4007/annals.2006.164.51. 6

[31] M. Grötschel, L. Lovasz, A. Schrijver, The ellipsoid method and its consequences in combinatorial optimization, Combinatorica 1 (2) (1981) 169–198. doi:10.1007/BF02579273. 6

[32] M. Chudnovsky, G. Cornuéjols, X. Liu, P. D. Seymour, K. Vušković, Recognizing Berge graphs, Combinatorica 25 (2) (2005) 143–186. doi:10.1007/s00493-005-0012-8. 6

[33] R. Takhanov, A Dichotomy Theorem for the General Minimum Cost Homomorphism Problem, in: Proceedings of the 27th International Symposium on Theoretical Aspects of Computer Science (STACS'10), 2010, pp. 657–668. doi:10.4230/LIPIcs.STACS.2010.2493. 4, 6

[34] T. Jebara, MAP Estimation, Message Passing, and Perfect Graphs, in: Proceedings of the Twenty-Fifth International Conference on Uncertainty in Artificial Intelligence (UAI), 2009, pp. 258–267. Available from: http://www.cs.mcgill.ca/~uai2009/papers/UAI2009_0098_c9e0cfea5b7aad26ceef02e3cef44909.pdf. 6

[35] P. David, Using Pivot Consistency to Decompose and Solve Functional CSPs, Journal Artificial Intelligence Research 2 (1995) 447–474. doi:10.1613/jair.167. 7

[36] M. Garey, D. Johnson, Computers and Intractability: A Guide to the Theory of NP-Completeness, W.H. Freeman, 1979. 10, 11, 19

[37] C. Papadimitriou, Computational Complexity, Addison-Wesley, 1994. 11, 19

[38] J.-C. Régin, A filtering algorithm for constraints of difference in CSPs, in: Proceedings of the 12th National Conference on AI (AAAI'94), Vol. 1, 1994, pp. 362–367. Available from: http://www.aaai.org/Papers/AAAI/1994/AAAI94-055.pdf. 12

[39] T. Petit, J.-C. Régin, C. Bessière, Specific filtering algorithms for over-constrained problems, in: Principles and Practice of Constraint Programming (CP'01), Vol. 2239 of Lecture Notes in Computer Science, Springer, 2001, pp. 451–463. doi:10.1007/3-540-45578-7_31. 12

[40] W. J. van Hoeve, G. Pesant, L.-M. Rousseau, On global warming: Flow-based soft global constraints, Journal of Heuristics 12 (4-5) (2006) 347–373. doi:10.1007/3-540-45578-7_31. 12, 15

[41] W. Horn, Minimizing average flow time with parallel machines, Operations Research 21 (3) (1973) 846–847. Available from: http://www.jstor.org/stable/169392. 13

[42] J. L. Bruno, E. G. C. Jr., R. Sethi, Scheduling independent tasks to reduce mean finishing time, Communications of the ACM 17 (7) (1974) 382–387. doi:10.1145/361011.361064. 13

[43] J.-C. Régin, Cost-based arc consistency for global cardinality constraints, Constraints 7 (3-4) (2002) 387–405. doi:10.1023/A:1020506526052. 15

[44] J. H.-M. Lee, K. L. Leung, Towards efficient consistency enforcement for global constraints in weighted constraint satisfaction, in: Proceedings of the 21st International Joint Conference on Artificial Intelligence (IJCAI'09), 2009, pp. 559–565. Available from: http://ijcai.org/papers09/Papers/IJCAI09-099.pdf. 15

[45] R. Ahuja, T. Magnanti, J. Orlin, Network Flows: Theory, Algorithms, and Applications, Prentice Hall/Pearson, 2005. 15, 16, 18

[46] A. Schrijver, Combinatorial Optimization: Polyhedra and Efficiency, Vol. 24 of Algorithms and Combinatorics, Springer, 2003. 16, 18

[47] T. Schaefer, The Complexity of Satisfiability Problems, in: Proceedings of the 10th Annual ACM Symposium on Theory of Computing (STOC'78), 1978, pp. 216–226. doi:10.1145/800133.804350. 19

[48] J.-C. Régin, Generalized Arc Consistency for Global Cardinality Constraint, in: Proceedings of the 13th National Conference on AI (AAAI'96), Vol. 1, 1996, pp. 209–215. Available from: http://www.aaai.org/Papers/AAAI/1996/AAAI96-031.pdf.